\documentclass[10pt,journal,compsoc]{IEEEtran}

\ifCLASSOPTIONcompsoc
  \usepackage[nocompress]{cite}
\else
  \usepackage{cite}
\fi
\ifCLASSINFOpdf
   \usepackage[pdftex]{graphicx}
   \DeclareGraphicsExtensions{.pdf,.jpeg,.png}
\else
\fi
\usepackage{amsmath,amsthm,amssymb,amsfonts,bbm,mathrsfs}
\newtheorem{theorem}{Theorem}

\newtheorem*{proof*}{Proof}
\newtheorem*{remark*}{Remark}

\usepackage{algorithmic,algorithm}

\usepackage{array}
\usepackage{multirow,multicol}
\usepackage{color,xcolor}
\usepackage{setspace}
\usepackage{caption,subcaption}
\usepackage{booktabs}
\usepackage{diagbox}

\begin{document}
%
\title{A Unified Framework for Coupled Tensor Completion}
%
%
%
%

\author{Huyan~Huang,~
Yipeng~Liu,~\IEEEmembership{Senior~Member,~IEEE},
Ce~Zhu,~\IEEEmembership{Fellow,~IEEE}
\IEEEcompsocitemizethanks{\IEEEcompsocthanksitem All the authors are with School of Information and Communication Engineering, University of Electronic Science and Technology of China (UESTC), Chengdu, 611731, China.\protect\\
E-mail: huyanhuang@gmail.com (or 201821011535@std.uestc.edu.cn), yipengliu@uestc.edu.cn, eczhu@uestc.edu.cn}
\thanks{This research is supported by National Natural Science Foundation of China (NSFC, No. 61602091, No. 61571102) and the Sichuan Science and Technology program (No. 2019YFH0008, No. 2018JY0035). The corresponding author is Yipeng Liu.}
}

%
%

\markboth{Journal of \LaTeX\ Class Files,~Vol.~XX, No.~X, Month~Year}%
{Shell \MakeLowercase{\textit{et al.}}: Bare Advanced Demo of IEEEtran.cls for IEEE Computer Society Journals}
%



\IEEEtitleabstractindextext{%
\begin{abstract}
Coupled tensor decomposition reveals the joint data structure by incorporating priori knowledge that come from the latent coupled factors. The tensor ring (TR) decomposition is invariant under the permutation of tensors with different mode properties, which ensures the uniformity of decomposed factors and mode attributes. The TR has powerful expression ability and achieves success in some multi-dimensional data processing applications. To let coupled tensors help each other for missing component estimation, in this paper we utilize TR for coupled completion by sharing parts of the latent factors. The optimization model for coupled TR completion is developed with a novel Frobenius norm. It is solved by the block coordinate descent algorithm which efficiently solves a series of quadratic problems resulted from sampling pattern. The excess risk bound for this optimization model shows the theoretical performance enhancement in comparison with other coupled nuclear norm based methods. The proposed method is validated on numerical experiments on synthetic data, and experimental results on real-world data demonstrate its superiority over the state-of-the-art methods in terms of recovery accuracy.
\end{abstract}

}

\maketitle

\IEEEdisplaynontitleabstractindextext

%
\IEEEpeerreviewmaketitle

\ifCLASSOPTIONcompsoc
\IEEEraisesectionheading{\section{Introduction}\label{section_introduction}}
\else
\section{Introduction}
\label{section_introduction}
\fi

%
%
%
%
\IEEEPARstart{T}{ensor} is a natural form to represent multi-dimensional data. The multi-way data processing performance can be effectively enhanced by tensor based techniques in comparison with the matrix counterparts \cite{cichocki2015tensor}. For instance, a color image can be regarded as a $3$-order tensor with two spatial modes and one channel mode, and tensor method can better exploit the coherence in all the modes simultaneously \cite{liu2012tensor, long2019low}.

\begin{figure}[htbp]
\centering
\includegraphics[scale=0.4]{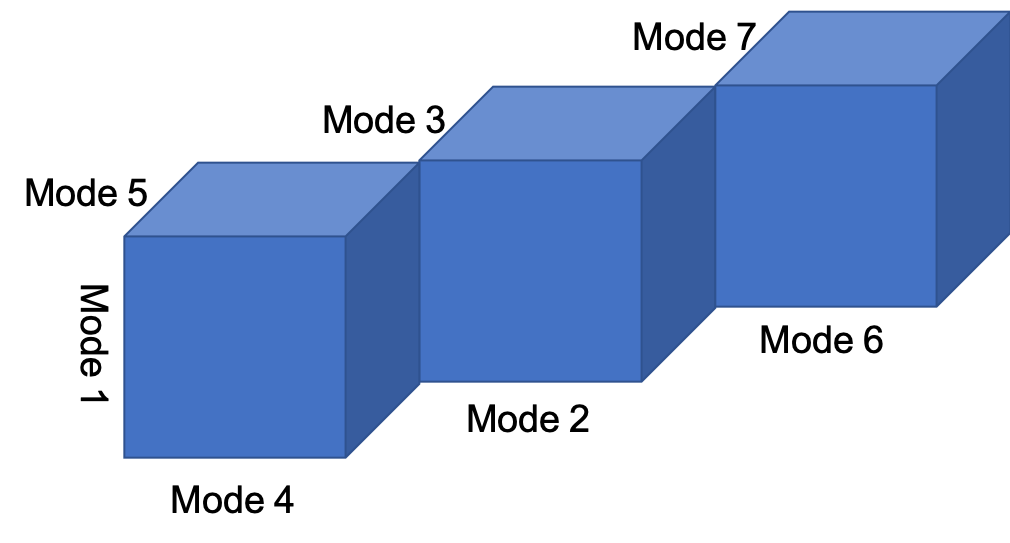}
\caption{Illustration of three coupled tensors on  mode-$1$.}
\label{illustration1}
\end{figure}

As it shows in Fig. \ref{illustration1}, multi-way data can be generated from different sources but share some same modes, and coupled tensor is a good representation for these data. These kinds of coupled tensors widely exist in bioinformatics \cite{gandy2011tensor, bazerque2013rank}, recommendation system \cite{symeonidis2016matrix}, link prediction \cite{liu2014factor, ermics2015link} and chemometrics \cite{narita2012tensor}.
During the acquisition and transmission, multi-way data can be partly corrupted, and tensor completion can recover the missing entries by low-rank approximation based on various tensor decompositions \cite{long2019low}. Coupled tensor decomposition gives an equivalent representation of multi-way data by a set of small factors, and parts of the factors are shared for coupled signals \cite{yilmaz2011generalised}. The corresponding coupled tensor completion can achieve better performance than the individual one by further exploiting the latent structures on coupled modes, which indicates that the freedom degree of the coupled system is decreased.

The completion methods are mainly divided into two categories. One is the convex method based on the optimization of low rank inducing nuclear norms, and the other one is the non-convex method based on the optimization of latent factors with pre-defined tensor rank. Most of current coupled tensor completion methods are based on CANDECOMP/PARAFAC (CP) decomposition and Tucker (TK) decomposition \cite{acar2011all, acar2013structure, acar2014structure, sorber2015structured}. By generalizing the singular value decomposition (SVD) for matrices, the CP decomposition factorizes a $D$-order tensor into a linear combination of $D$ rank-$1$ tensors, resulting in $DIR$ parameters, where $I$ is the dimensional size and $R$ is the CP rank \cite{zhou2019tensor}. The TK decomposition gives a core tensor mode-multiplied by a number of matrices  \cite{liu2020low}.

\begin{figure*}[htbp]
\centering
\begin{subfigure}[t]{0.48\textwidth}
\centering
\includegraphics[scale=0.42]{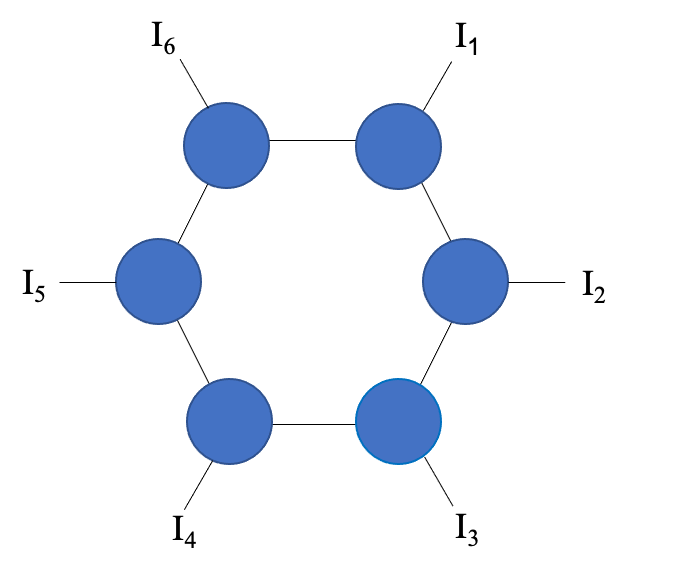}
\subcaption{A graphical TR representation for a $6$-order tensor.}
\end{subfigure}
\;
\begin{subfigure}[t]{0.48\textwidth}
\centering
\includegraphics[scale=0.42]{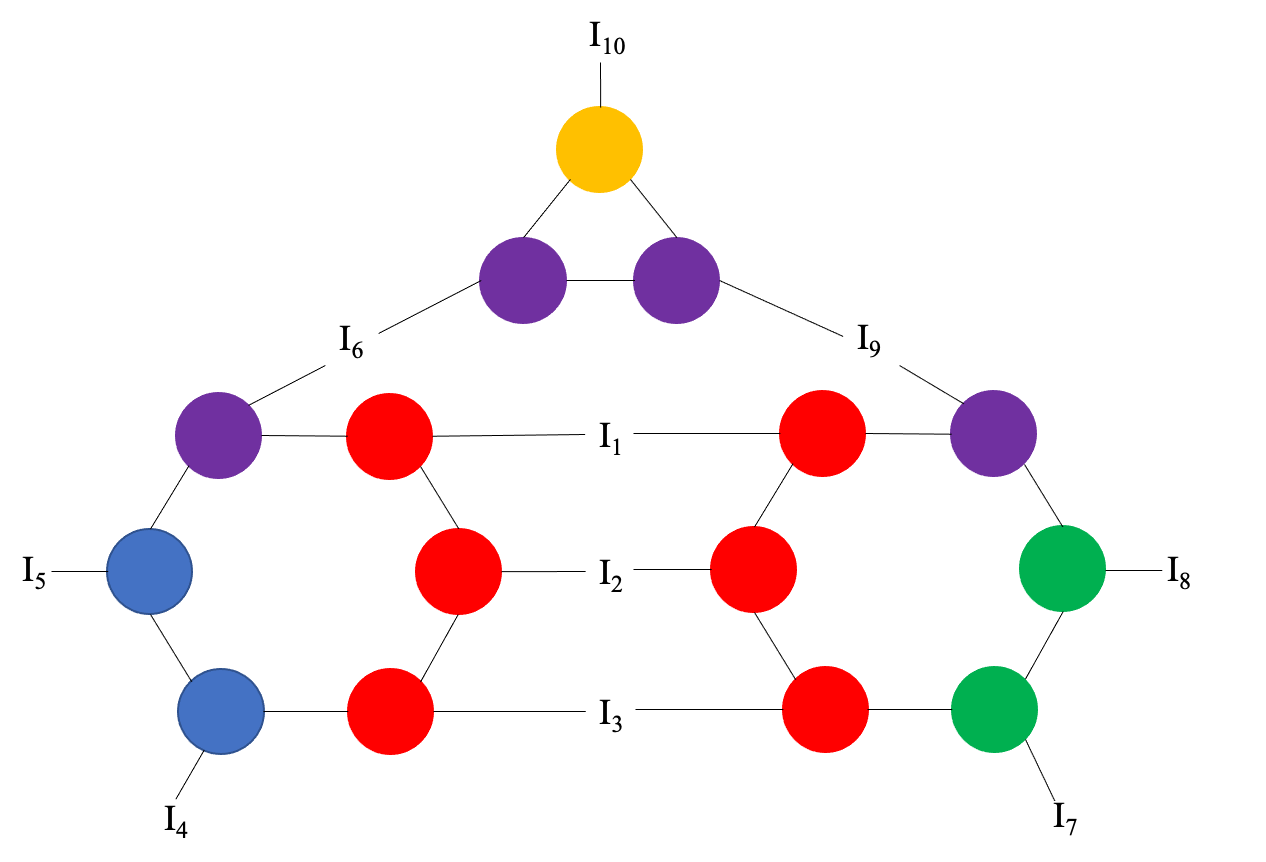}
\subcaption{A graphical coupled TR representation for $3$ tensors. The disks with the same color mean they have the same factors respectively, e.g., mode $1$, $2$ and $3$ are shared by the first two tensors and mode $6$ and $9$ are shared by three tensors.}
\end{subfigure}
\caption{Illustration of the coupled TR decomposition.}
\label{illustration2}
\end{figure*}

The recently proposed tensor ring (TR) decomposition represents a $D$-order tensor with cyclically contracted $3$-order tensor factors of size $R\times I\times R$ by using the matrix product state expression \cite{huang2020provable}. As shown in Fig. \ref{illustration2}(a), it has $DIR^2$ parameters, where $\left[R; \cdots ; R\right]$ is the TR rank. The TR decomposition allows a cyclical shift of factors due to the nature of trace operator, and reordering tensor dimension makes no difference to the decomposition. As a quantum-inspired decomposition, it outperforms CP decomposition and TK decomposition due to its powerful representation ability in many applications \cite{bengua2017efficient, wang2017efficient}. Though the TR rank is a vector, it is approximately effective to let all components have the same value \cite{wang2017efficient}, which alleviates the burden for tuning parameters. In \cite{xu2020hyperspectral}, TR is used for coupled tensor fusion with different dimensional sizes, i.e., the fusion of multispectral images and hyperspectral images. There is no TR completion with their factors directly coupled, as given in Fig. \ref{illustration2}(b).

To the best of our knowledge, this is the first attempt to use TR for coupled tensor completion. Given pre-defined TR ranks, the coupled TR decomposition can be formulated as the optimization with respect to factors such that the deviation of the approximation from the given tensors is minimized. In this paper, we propose the low rank coupled TR completion (CTRC), which can be regarded as the coupled TR decomposition for incomplete tensors. The block coordinate descent (BCD) algorithm is used to solve the problem. The computation and storage complexity are analyzed. In numerical experiments, the proposed CTRC-BCD is tested on synthetic data to verify the theoretical analysis. The proposed method is benchmarked on the user-centered collaborative location and activity filtering (UCLAF), short-wave near-infrared spectrum (SW-NIR) and electronic nose licorice datasets. The experimental results demonstrate the proposed method outperforms the state-of-the-art ones.The main contributions of this paper are as follows:
\begin{enumerate}
\item We propose a coupling model based on TR where arbitrary tensors are allowed to be coupled by sharing any parts of their TR factors (a scenario of two coupled tensors is shown in Fig. \ref{illustration2}). The coupled TR decomposition calculates its factors by solving a non-linear fitting problem that aims to minimize the squared error of the difference between the given tensors and their estimates. Either part or whole of the entries in the coupled factors are shared in the proposed model. The TR ranks need to be pre-defined.

\item The block coordinate descent algorithm is used to solve the coupled TR completion problem. It alternately solves a series of quadratic forms with respect to the latent factors. The Hessian matrix depends on its corresponding sampling pattern, which results in an efficient updating scheme.

\item With the newly defined coupled Frobenius norm (F-norm) for coupled tensors, we derive an excess risk bound using the recently proposed permutational Rademacher complexity \cite{tolstikhin2015permutational} with a modern mathematical tool called brackets method \cite{gonzalez2010method, gonzalez2017extension}. The conclusion indicates that coupled tensors possess a lower sampling bound which is lower than each individual tensor's sampling bound.
\end{enumerate}

The rests of this paper are organized as follows. In Section \ref{section_notation}, we introduce basic notations and preliminaries of tensor, TR decomposition and its relevant operations. In Section \ref{section_algorithm}, we state the CTRC problem and propose our algorithm, along with the algorithmic complexity. We provide an excess risk bound for the coupled TR F-norm model in Section \ref{section_bound}. In Section \ref{section_experiment}, we perform a series of numerical experiments to compare the proposed method with the existing ones. Finally we conclude our work in Section \ref{section_conclusion}. 


\section{Notations and Preliminaries}
\label{section_notation}

\subsection{Notations}

Throughout the paper, a scalar, a vector, a matrix and a tensor are denoted by a normal letter, a boldfaced lower-case letter, a boldfaced upper-case letter and a calligraphic letter, respectively. For instance, a $D$-order tensor is denoted as $\mathcal{X} \in \mathbb{R}^{I_1 \times \dotsm \times I_D}$, where $I_d$ is the dimensional size for $d$-th mode, $d=1, \cdots, D$.

The Frobenius norm of $\mathcal{X}$ is defined as the squared root of the inner product of two tensors:
\begin{equation}
\lVert \mathcal{X} \rVert_\mathrm{F}=\sqrt{\langle \mathcal{X},\mathcal{X} \rangle}=\sqrt{\sum_{i_1=1}^{I_1}\cdots \sum_{i_D=1}^{I_D}{x_{i_1\cdots i_D}^2}}.
\end{equation}
Projection $\operatorname{P}_{\mathbb{O}}: \mathbb{R}^{I_1 \times \dotsm \times I_D} \mapsto \mathbb{R}^m$ projects a tensor onto the support (observation) set $\mathbb{O}$, where
\begin{equation}
\mathbb{O}:=\left\{\left(i_1,\dotsc,i_D\right)|\text{ entry on }\left(i_1,\dotsc,i_D\right)\text{ is observed}\right\},
\end{equation}
where $\operatorname{P}_{\mathbb{O}}\left(\mathcal{X}\right)$ represents the observed entries in $\mathcal{X}$ as
\begin{equation}
\operatorname{P}_{\mathbb{O}}\left(\mathcal{X}\right)_{i_1\dotsm i_D}=
\left\{
\begin{aligned}
x_{i_1\dotsm i_D} \;\;\;& \left(i_1,\dotsc,i_D\right)\in \mathbb{O} \\
0 \;\;\;& \left(i_1,\dotsc,i_D\right)\notin \mathbb{O}
\end{aligned}
\right..
\end{equation}

The Hadamard product $\circledast$ is an element-wise product. For $D$-th order tensors $\mathcal{X}$ and $\mathcal{Y}$, the representation is
\begin{equation}
\left(\mathcal{X} \circledast \mathcal{Y} \right)_{i_1\dotsm i_D}=x_{i_1\dotsm i_D}\cdot y_{i_1\dotsm i_D}.
\end{equation}

The $d$-shifting $H$-unfolding yields a matrix $\mathbf{X}_{\left\{d,H\right\}}$ by permuting $\mathcal{X}$ with order $\left[d,\dotsc,D, 1,\dotsc,d-1\right]$ and unfolding along its first $H$ dimensions.

As a natural extension of the traditional F-norm, the couple F-norm for coupled tensors can be defined as follows:
\begin{align}
\left\| \mathcal{X},\mathcal{Y} \right\|_{\mathrm{CF}}\triangleq \sqrt{\left\| \mathcal{X} \right\|^2_{\mathrm{F}}+\left\| \mathcal{Y} \right\|^2_{\mathrm{F}}},
\end{align}
whose dual norm is itself. The newly defined coupled F-norm can be used as an accuracy measure for coupled tensor completion.

\subsection{Preliminaries of tensor ring decomposition}

The non-canonical TR decomposition factorizes $\mathcal{X} \in \mathbb{R}^{I_1\times \dotsm \times I_D} $ into $D$ cyclically contracted $3$-order tensors as follows \cite{orus2014practical}:
\begin{align}
\mathcal{X}\left(i_1,\dotsc,i_D\right)=\operatorname{tr}\left( \mathcal{U}^{\left(1\right)}\left(:,i_1,:\right)\dotsm\mathcal{U}^{\left(D\right)}\left(:,i_D,:\right) \right),
\label{TR decomposition}
\end{align}
where $\mathcal{U}^{\left(d\right)}\in \mathbb{R}^{R_d\times I_d\times R_{d+1}}$ and $\operatorname{tr}\left(\cdot\right)$ is the trace norm.

Two methods are introduced for TR decomposition in \cite{zhao2017learning}. The first one is based on the density matrix renormalization group \cite{oseledets2011tensor}. It firstly reshapes $\mathcal{X}$ into $\mathbf{X}_{\left\{1,1\right\}}$ and applies SVD to derive $\mathbf{X}_{\left\{1,1\right\}}=\mathbf{U}\boldsymbol{\Sigma}\mathbf{V}$. It then reshapes $\mathbf{U}$ as the first TR factor and applies SVD to $\boldsymbol{\Sigma}\mathbf{V}$.  $D-1$ SVDs are performed afterwards. This method does not per-define TR rank and performs fast. The second method alternatively optimizes one of the TR factors while keeping the others fixed. It repeatedly performs the optimization until the relative change $\left\| \mathcal{X}^k-\mathcal{X}^{k-1} \right\|_\mathrm{F}/\left\| \mathcal{X}^{k-1} \right\|_\mathrm{F}$ or the relative error $\left\| \mathcal{X}^k-\mathcal{X} \right\|_\mathrm{F}/\left\| \mathcal{X} \right\|_\mathrm{F}$ decreases below a certain pre-defined threshold. This method requires a pre-defined TR rank which affects the performance, and it is slower than the first one.

\section{Method}
\label{section_algorithm}

We use $\mathfrak{R}$ to represent the TR computation, and assume the first $L$ factors of two TRs are coupled without loss of generality. Operator $\mathfrak{R}\left(\cdot\right)$ means the TR contraction which yields a tensor given a set of TR factors. Supposing the TR factors of $\mathfrak{R}_1$ and $\mathfrak{R}_2$ are 
$\left\{ \mathcal{U} \right\}=\left\{ \mathcal{U}^{\left(1\right)},\dotsc,\mathcal{U}^{\left(D_1\right)} \right\}$  and $\left\{ \mathcal{V} \right\}=\left\{ \mathcal{V}^{\left(1\right)},\dotsc,\mathcal{V}^{\left(D_2\right)} \right\}$, respectively, we have $\mathfrak{R}\left( \left\{ \mathcal{U} \right\} \right)\in \mathbb{R}^{I_1\times \dotsm \times I_{D_1}}$ and $\mathfrak{R}\left( \left\{ \mathcal{V} \right\} \right)\in \mathbb{R}^{{I'}_1\times \dotsm \times {I'}_{D_2}}$.

Given the coupled measurements $\mathcal{T}_1$ and $\mathcal{T}_2$, the optimization model for CTRC can be formulated as follows:
\begin{equation}
\begin{aligned}
\min_{\left\{ \mathcal{U} \right\},\left\{ \mathcal{V} \right\}}\; & \frac{1}{2}\lVert \operatorname{P}_{\mathbb{O}_1}\left( \mathfrak{R}\left(\left\{ \mathcal{U} \right\}\right) \right)-\operatorname{P}_{\mathbb{O}_1}\left( \mathcal{T}_1 \right) \rVert^2_2+ \\
& \frac{1}{2}\lVert \operatorname{P}_{\mathbb{O}_2}\left( \mathfrak{R}\left(\left\{ \mathcal{V} \right\}\right) \right)-\operatorname{P}_{\mathbb{O}_2}\left( \mathcal{T}_2 \right) \rVert^2_2 \\
\text{s. t.}\;\;\; & \mathcal{U}^{\left(d\right)}\left(1:\Gamma_d,:,1:\Gamma_{d+1}\right)=\mathcal{V}^{\left(d\right)}\left(1:\Gamma_d,:,1:\Gamma_{d+1}\right) \\
& d=1,\dotsc,L,
\end{aligned}
\label{model-coupled tensor ring completion}
\end{equation}
where $\Gamma_d\in \left[1,\min\left\{R_d,{R'}_d\right\}\right]$, $d=1,\dotsc,L$ are the coupled distances, in which $\left[R_1; \cdots; R_{D_1}\right]$ and $\left[{R'}_1; \cdots; {R'}_{D_2}\right]$ are the TR ranks of $\mathfrak{R}\left( \left\{ \mathcal{U} \right\} \right)$ and $\mathfrak{R}\left( \left\{ \mathcal{V} \right\} \right)$, respectively.

\subsection{Algorithm}

\begin{figure*}[htbp]
\centering
\begin{subfigure}[t]{0.48\textwidth}
\centering
\includegraphics[scale=0.42]{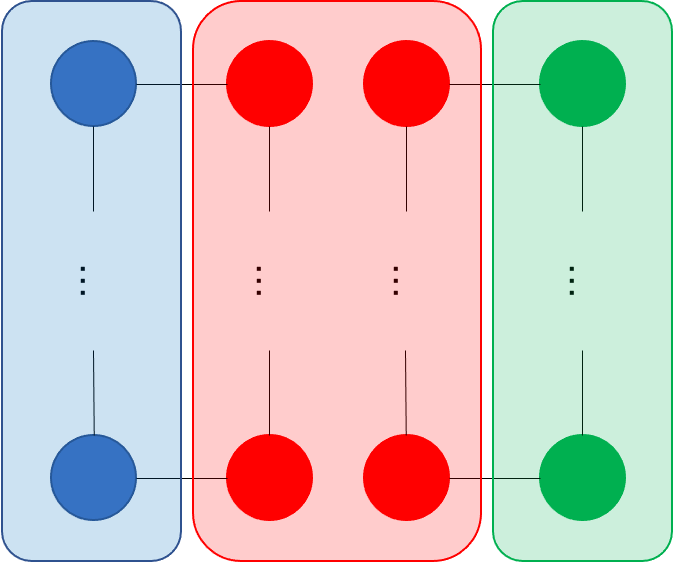}
\subcaption{Step $1$: divide the variables into $3$ parts, the blue and the green factors are uncoupled ones for each TR, and the red factors are coupled factors.}
\label{step1}
\end{subfigure}
\;
\begin{subfigure}[t]{0.48\textwidth}
\centering
\includegraphics[scale=0.42]{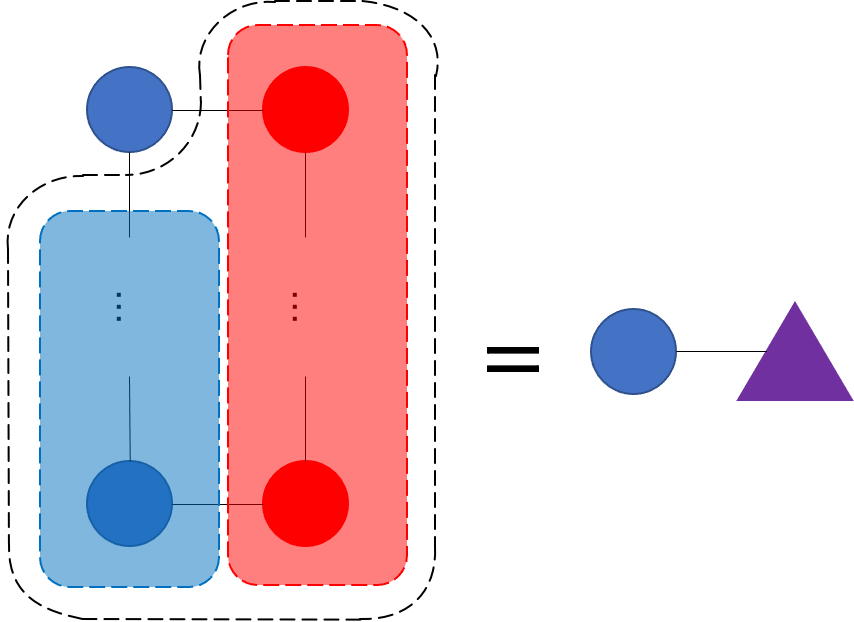}
\subcaption{Step $2$: consider the uncoupled factors of the first TR. The coupled factors are contracted first, then sling out one uncoupled factor per round and contract the others, in the next all factors except the single uncoupled one are contracted,  which we use a purple triangle to represent. The problem now becomes a matrix equation with linear sampling.}
\label{step2}
\end{subfigure}

\begin{subfigure}[t]{0.48\textwidth}
\centering
\includegraphics[scale=0.42]{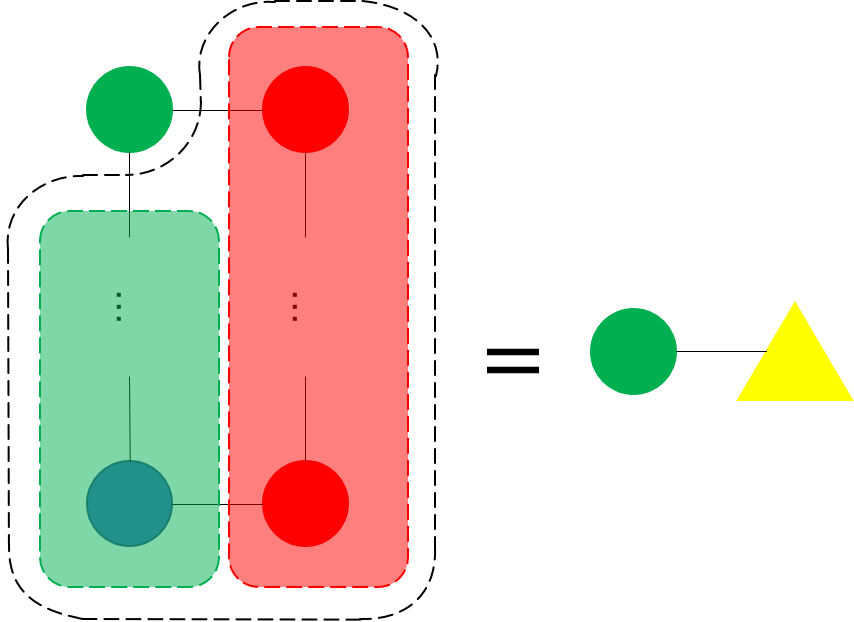}
\subcaption{Step $3$: consider the uncoupled factors of the second TR, for which the optimization is the same as the first TR.}
\label{step3}
\end{subfigure}
\;
\begin{subfigure}[t]{0.48\textwidth}
\centering
\includegraphics[scale=0.32]{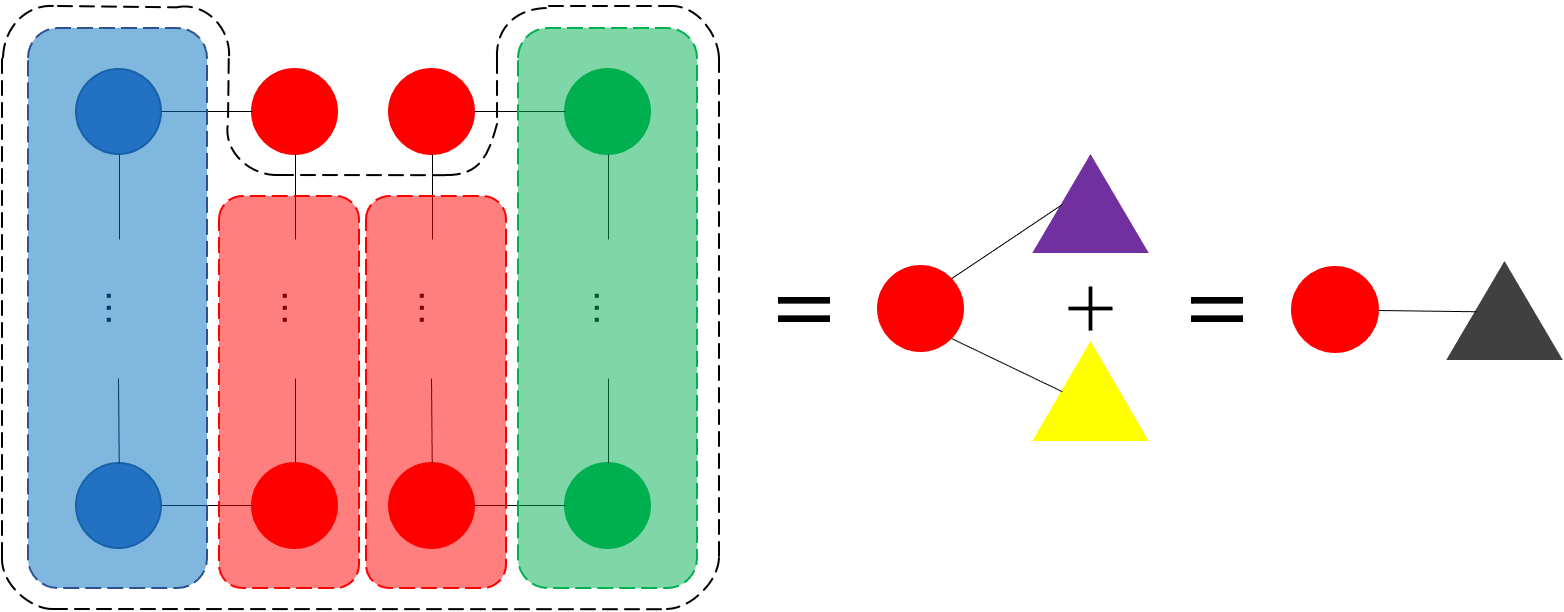}
\subcaption{Step $4$: consider the coupled factors. The optimization process is similar to Step $2$, but the contracted factors need to be add up just like the purple triangle and the yellow triangle shown above.}
\label{step4}
\end{subfigure}
\caption{A graphic illustration for the optimization which contains $4$ steps.}
\label{steps}
\end{figure*}

To solve problem (\ref{model-coupled tensor ring completion}), we use the block coordinate descent algorithm \cite{xu2013block}. Specifically, it alternately optimizes the block variable $\mathcal{U}^{\left(d_1\right)}$ (or $\mathcal{V}^{\left(d_2\right)}$) while keeping others fixed, thus problem (\ref{model-coupled tensor ring completion}) is decomposed into $D_1+D_2-L$ sub-problems. Fig. \ref{steps} shows $4$ steps for optimization. 

\subsubsection{Update of the uncoupled factors of first TR}

we reformulate the problem (\ref{model-coupled tensor ring completion}) into
\begin{equation}
\begin{aligned}
\min_{\substack{\mathcal{U}^{\left(d_1\right)} \\ d_1=L+1,\dotsc,D_1}}\; & \frac{1}{2}\lVert \operatorname{P}_{\mathbb{O}_1}\left( \mathfrak{R}\left(\left\{ \mathcal{U} \right\}\right) \right)-\operatorname{P}_{\mathbb{O}_1}\left( \mathcal{T}_1 \right) \rVert^2_2
\end{aligned}
\label{model-tensor ring1}
\end{equation}
along with substitution of $\mathcal{W}_1\circledast \left( \cdot \right)$ for $\operatorname{P}_{\mathbb{O}_1}\left( \cdot \right)$, let $\mathbf{A}_{d_1}\in \mathbb{R}^{I_{d_1}\times R_{d_1}R_{d_1+1}}$, $\mathbf{B}_{d_1}\in \mathbb{R}^{R_{d_1}R_{d_1+1}\times J_d}$ and $\mathbf{C}_{d_1}\in \mathbb{R}^{I_{d_1}\times J_{d_1}}$ be the unfoldings of $\mathcal{U}^{\left(d_1\right)}$, $\mathcal{B}_{d_1}$ and $\mathcal{T}_1$, respectively, where $\mathcal{B}_{d_1}$ is computed by contracting all the $D_1$ TR factors of $\mathfrak{R}_1$ except the $d_1$-th factor. Then we can get an equivalent form of (\ref{model-tensor ring1}) as follows:
\begin{equation}
\begin{aligned}
\min_{\substack{\mathbf{A}_{d_1} \\ d_1=L+1,\dotsc,D_1}}\; & \frac{1}{2}\lVert \mathbf{W}_{\left\{d_1,1\right\}}\circledast \mathbf{A}_{d_1}\mathbf{B}_{d_1}-\mathbf{W}_{1\left\{d_1,1\right\}}\circledast \mathbf{C}_{d_1} \rVert^2_{\mathrm{F}}.
\end{aligned}
\label{model-tensor ring1_d}
\end{equation}

Defining $\mathbf{w}^{\left(d_1\right)}_{i_{d_1}}=\mathbf{W}_{\left\{d_1,1\right\}}\left(i_{d_1},:\right)$ and the permutation matrix $\mathbf{Q}^{\left(d_1\right)}_{i_{d_1}}=\mathbf{e}_{\mathbb{S}^{\left(d_1\right)}_{i_{d_1}}}\in \mathbb{R}^{I_{\neq d_1}\times \lVert \mathbf{w}^{\left(d_1\right)}_{i_{d_1}} \rVert_0}$, where $\mathbf{e}_k$ is a vector of length $I_{\neq d_1}=\prod^{D_1}_{t=1}I_t/I_{d_1}$ whose values are all zero but one in the $k$-th entry, $k\in \mathbb{S}^{\left(d_1\right)}_{i_{d_1}}$ and $\mathbb{S}^{\left(d_1\right)}_{i_{d_1}}=\left\{j_{d_1} | \mathbf{w}^{\left(d_1\right)}_{i_{d_1}}\left(j_{d_1}\right)=1 \right\}$.

Note that the $d_1$-th sub-problem in (\ref{model-tensor ring1_d}) can be divided into $I_{d_1}$ sub-sub-problems, in which the row vectors $\mathbf{a}^{\left(d_1\right)}_{i_{d_1}}=\mathbf{A}_{d_1}\left(i_{d_1},:\right)$ are taken as the block variables. Reformulating the $i_{d_1}$-th sub-sub-problem in the quadratic form and calculating its first-order derivative, we have the optimal solution of the form (please refer to Appendix A for detail)
\begin{align}
\mathbf{a}^{\left(d_1\right)^*}_{i_{d_1}}=-\mathbf{g}^{\left(d_1\right)}_{i_{d_1}}\mathbf{H}^{\left(d_1\right)^{\dagger}}_{i_{d_1}},
\label{update1}
\end{align}
where $\dagger$ is the Moore-Penrose pseudo inverse, and
\begin{equation*}
\left\{
\begin{aligned}
& \mathbf{H}^{\left(d_1\right)}_{i_{d_1}}=\overline{\mathbf{B}}^{\left(d_1\right)}_{i_{d_1}}\overline{\mathbf{B}}^{{\left(d_1\right)}^{\mathrm{T}}}_{i_{d_1}},\; \mathbf{g}^{\left(d_1\right)}_{i_{d_1}}=-\overline{\mathbf{c}}^{\left(d_1\right)}_{i_{d_1}}\overline{\mathbf{B}}^{{\left(d_1\right)}^{\mathrm{T}}}_{i_{d_1}} \\
& \overline{\mathbf{c}}^{\left(d_1\right)}_{i_{d_1}}=\mathbf{c}^{\left(d_1\right)}_{i_{d_1}}\mathbf{Q}^{\left(d_1\right)}_{i_{d_1}},\; \overline{\mathbf{B}}^{\left(d_1\right)}_{i_{d_1}}=\mathbf{B}^{\left(d_1\right)}\mathbf{Q}^{\left(d_1\right)}_{i_{d_1}}
\end{aligned}
\right..
\end{equation*}

The TR factor $\mathcal{U}^{\left(d_1\right)}$ is optimized by performing (\ref{update1}) $I_{d_1}$ times to solve the $d_1$-th sub-problem of (\ref{model-tensor ring1_d}). The uncoupled TR factors of $\mathfrak{R}_1$ are updated by optimizing all $D_1-L$ factors.

\subsubsection{Update of the uncoupled factors of the second TR}

This optimization is similar to the update of uncoupled TR factors of $\mathfrak{R}_1$ by (\ref{update1}), and we neglect the deduction and just give the solution as follows:
\begin{align}
\mathbf{a'}^{\left(d\right)^*}_{{i'}_{d_2}}-\mathbf{g'}^{\left(d_2\right)}_{{i'}_{d_2}}\mathbf{H'}^{\left(d_2\right)^{\dagger}}_{{i'}_{d_2}},
\label{update2}
\end{align}
where $\mathbf{H'}^{\left(d_2\right)}_{{i'}_{d_2}}=\overline{\mathbf{B'}}^{\left(d_2\right)}_{{i'}_{d_2}}\overline{\mathbf{B'}}^{\left(d_2\right)^{\mathrm{T}}}_{{i'}_{d_2}}$, $\mathbf{g'}^{\left(d_2\right)}_{{i'}_{d_2}}=-\overline{\mathbf{c'}}^{\left(d_2\right)}_{{i'}_{d_2}}\mathbf{B'}_d^{\mathrm{T}}$, $z^{\left(d_2\right)}_{{i'}_{d_2}}=\overline{\mathbf{c'}}^{\left(d_2\right)}_{{i'}_{d_2}}\overline{\mathbf{c'}}_{{i'}_{d_2}}^{{\left(d_2\right)}^{\mathrm{T}}}$, and the symbols with superscript $'$ means that the corresponding terms are derived from computation of the second TR.

\subsubsection{Update of the coupled factors of two TRs}

we can rewrite the problem (\ref{model-coupled tensor ring completion}) as follows:
\begin{equation}
\begin{aligned}
\min_{\substack{\mathcal{U}^{\left(d\right)},\mathcal{V}^{\left(d\right)}\\ d=1,\dotsc,L}}\; & \frac{1}{2}\lVert \operatorname{P}_{\mathbb{O}_1}\left( \mathfrak{R}\left(\left\{ \mathcal{U} \right\}\right) \right)-\operatorname{P}_{\mathbb{O}_1}\left( \mathcal{T}_1 \right) \rVert^2_2+ \\
& \frac{1}{2}\lVert \operatorname{P}_{\mathbb{O}_2}\left( \mathfrak{R}\left(\left\{ \mathcal{V} \right\}\right) \right)-\operatorname{P}_{\mathbb{O}_2}\left( \mathcal{T}_2 \right) \rVert^2_2 \\
\text{s. t.}\quad & \mathcal{U}^{\left(d\right)}\left(1:\Gamma_d,:,1:\Gamma_{d+1}\right)= \\
& \mathcal{V}^{\left(d\right)}\left(1:\Gamma_d,:,1:\Gamma_{d+1}\right),\; d=1,\dotsc,L.
\end{aligned}
\label{model-tensor ring3}
\end{equation}
Let $\mathbf{A'}_d\in \mathbb{R}^{{I'}_d\times R_dR_{d+1}}$ be the unfolding of $\mathcal{V}^{\left(d\right)}$, $\mathbf{C'}_d\in \mathbb{R}^{{I'}_d\times {J'}_d}$ be the $\left\{d,1\right\}$ unfolding of $\mathcal{T}_2$ and $\mathcal{W'}$ be the tensor form of $\operatorname{P}_{\mathbb{O}_2}$, and
\begin{align*}
\mathbb{C}_d=\{& 1,\dotsc,\Gamma_{d+1},R_{d+1}+1,\dotsc,R_{d+1}+\Gamma_{d+1}, \\
& \dotsc, \\
& \Gamma_{d}R_{d+1}+1,\dotsc,\Gamma_{d}R_{d+1}+\Gamma_{d+1} \},\; d=1,\dotsc,L, \\
\mathbb{C}'_d=\{& 1,\dotsc,\Gamma_{d+1},{R'}_{d+1}+1,\dotsc,{R'}_{d+1}+\Gamma_{d+1}, \\
& \dotsc, \\
& \Gamma_{d}{R'}_{d+1}+1,\dotsc,\Gamma_{d}{R'}_{d+1}+\Gamma_{d+1} \},\; d=1,\dotsc,L.
\end{align*}
An equivalence for (\ref{model-tensor ring3}) can be obtained as follows:
\begin{equation}
\begin{aligned}
\min_{\substack{\mathbf{A}_d,\mathbf{A'}_d\\ d=1,\dotsc,L}}\; & \frac{1}{2}\lVert \mathbf{W}_{\left\{d,1\right\}}\circledast \mathbf{A}_d\mathbf{B}_d-\mathbf{W}_{\left\{d,1\right\}}\circledast \mathbf{C}_d \rVert^2_{\mathrm{F}}+ \\
& \frac{1}{2}\lVert \mathbf{W'}_{\left\{d,1\right\}}\circledast \mathbf{A'}_d\mathbf{B'}_d-\mathbf{W'}_{\left\{d,1\right\}}\circledast \mathbf{C}'_d \rVert^2_{\mathrm{F}} \\
\text{s. t.}\quad & \mathbf{A}_d\left(:,\mathbb{C}_d\right)=\mathbf{A'}_d\left(:,\mathbb{C}'_d\right),\; d=1,\dotsc,L,
\end{aligned}
\label{model-tensor ring3_d}
\end{equation}
where the index sets $\mathbb{C}_d\in \mathbb{R}^{\Gamma_d\Gamma_{d+1}}$ and $\mathbb{C'}_d\in \mathbb{R}^{\Gamma_d\Gamma_{d+1}}$ indicate which columns are coupled in $\mathbf{A}_d$ and $\mathbf{A}'_d$, respectively.

The $i_d$-th sub-sub-problem of the $d$-th sub-problem of (\ref{model-tensor ring3_d}) can take in the similar way as it does for optimization of (\ref{model-tensor ring1_d}). We split the variables in coupled factors into three new blocks, which are defined as
\begin{equation*}
\left\{
\begin{aligned}
& \boldsymbol{\alpha}^{\left(d\right)}_{i_d}\triangleq \mathbf{A}_d\left(i_d,\mathbb{C}_d\right)=\mathbf{A'}_d\left(i_d,\mathbb{C'}_d\right) \\
& \boldsymbol{\beta}^{\left(d\right)}_{i_d}\triangleq \mathbf{A}_d\left(i_d,\left\{1,\dotsc,R_dR_{d+1}\right\}\backslash \mathbb{C}_d\right) \\
& \boldsymbol{\gamma}^{\left(d\right)}_{i_d}\triangleq \mathbf{A'}_d\left(i_d,\left\{1,\dotsc,{R'}_d{R'}_{d+1}\right\}\backslash \mathbb{C'}_d\right)
\end{aligned}
\right.,
\end{equation*}
and the equivalent relationship hold as follows:
\begin{equation*}
\left\{
\begin{aligned}
& \mathbf{A}_d\left(i_d,:\right)=\left[\boldsymbol{\alpha}^{\left(d\right)}_{i_d},\boldsymbol{\beta}^{\left(d\right)}_{i_d}\right]\mathbf{P}^{\mathrm{T}}_d \\
& \mathbf{A'}_d\left(i_d,:\right)=\left[\boldsymbol{\alpha}^{\left(d\right)}_{i_d},\boldsymbol{\gamma}^{\left(d\right)}_{i_d}\right]\mathbf{P'}^{\mathrm{T}}_d
\end{aligned}
\right.,
\end{equation*}
where
\begin{equation*}
\left\{
\begin{aligned}
& \mathbf{P}_d=\left[\mathbf{e}_{\mathbb{C}_d}; \mathbf{e}_{\left\{1,\dotsc,R_dR_{d+1}\right\}\backslash \mathbb{C}_d}\right] \\
& \mathbf{P'}_d=\left[\mathbf{e}_{\mathbb{C'}_d}; \mathbf{e}_{\left\{1,\dotsc,{R'}_d{R'}_{d+1}\right\}\backslash \mathbb{C'}_d}\right]
\end{aligned}
\right.
\end{equation*}
are permutation matrices. Accordingly, we can get 
\begin{equation}
\begin{aligned}
& \min_{\boldsymbol{\alpha}^{\left(d\right)}_{i_d},\boldsymbol{\beta}^{\left(d\right)}_{i_d},\boldsymbol{\gamma}^{\left(d\right)}_{i_d}} \frac{1}{2}\lVert \mathbf{w}^{\left(d\right)}_{i_d}\circledast \left[\boldsymbol{\alpha}^{\left(d\right)}_{i_d},\boldsymbol{\beta}^{\left(d\right)}_{i_d}\right]\mathbf{P}^{\mathrm{T}}_d\mathbf{B}_d-\mathbf{w}^{\left(d\right)}_{i_d}\circledast \mathbf{c}^{\left(d\right)}_{i_d} \rVert^2_2 \\
&\qquad +\frac{1}{2}\lVert \mathbf{w'}^{\left(d\right)}_{i_d}\circledast \left[\boldsymbol{\alpha}^{\left(d\right)}_{i_d},\boldsymbol{\gamma}^{\left(d\right)}_{i_d}\right]\mathbf{P'}^{\mathrm{T}}_d\mathbf{B'}_d-\mathbf{w'}^{\left(d\right)}_{i_d}\circledast \mathbf{c'}^{\left(d\right)}_{i_d} \rVert^2_2.
\end{aligned}
\label{model-tensor ring3_i_d}
\end{equation}

Letting $\widetilde{\mathbf{H}}^{\left(d\right)}_{i_d}=\mathbf{P}^{\mathrm{T}}_d\mathbf{H}^{\left(d\right)}_{i_d}\mathbf{P}_d$ and $\widetilde{\mathbf{H'}}^{\left(d\right)}_{i_d}=\mathbf{P'}^{\mathrm{T}}_d\mathbf{H'}^{\left(d\right)}_{i_d}\mathbf{P'}_d$, we reformulate the Hessian matrices in a block form
\begin{align*}
\widetilde{\mathbf{H}}^{\left(d\right)}_{i_d}\triangleq
\begin{bmatrix}
\widetilde{\mathbf{H}}_{i_d}^{\left(d\right)11} & \widetilde{\mathbf{H}}_{i_d}^{\left(d\right)12} \\
\widetilde{\mathbf{H}}_{i_d}^{\left(d\right)21} & \widetilde{\mathbf{H}}_{i_d}^{\left(d\right)22}
\end{bmatrix}
,\; 
\widetilde{\mathbf{H'}}^{\left(d\right)}_{i_d}\triangleq
\begin{bmatrix}
\widetilde{\mathbf{H'}}_{i_d}^{\left(d\right)11} & \widetilde{\mathbf{H'}}_{i_d}^{\left(d\right)12} \\
\widetilde{\mathbf{H'}}_{i_d}^{\left(d\right)21} & \widetilde{\mathbf{H'}}_{i_d}^{\left(d\right)22}
\end{bmatrix}
\end{align*}
such that
\begin{align*}
& \widetilde{\mathbf{H}}_{i_d}^{\left(d\right)11}\in \mathbb{R}^{\Gamma_d\Gamma_{d+1}\times \Gamma_d\Gamma_{d+1}}, \\
& \widetilde{\mathbf{H}}_{i_d}^{\left(d\right)12}\in \mathbb{R}^{\Gamma_d\Gamma_{d+1}\times R_dR_{d+1}-\Gamma_d\Gamma_{d+1}}, \\
& \widetilde{\mathbf{H}}_{i_d}^{\left(d\right)21}\in \mathbb{R}^{R_dR_{d+1}-\Gamma_d\Gamma_{d+1}\times \Gamma_d\Gamma_{d+1}}, \\ 
& \widetilde{\mathbf{H}}_{i_d}^{\left(d\right)22}\in \mathbb{R}^{R_dR_{d+1}-\Gamma_d\Gamma_{d+1}\times R_dR_{d+1}-\Gamma_d\Gamma_{d+1}}
\end{align*}
and the similar sizes hold for $\widetilde{\mathbf{H'}}_{i_d}^{\left(d\right)11}$, $\widetilde{\mathbf{H'}}_{i_d}^{\left(d\right)12}$, $\widetilde{\mathbf{H'}}_{i_d}^{\left(d\right)21}$ and $\widetilde{\mathbf{H'}}_{i_d}^{\left(d\right)22}$. Define
\begin{equation*}
\left\{
\begin{aligned}
& \mathbf{g}^{\left(d\right)}_{i_d}\triangleq -\overline{\mathbf{c}}^{\left(d\right)}_{i_d}\overline{\mathbf{B}}^{\left(d\right)^{\mathrm{T}}}_{i_d}\mathbf{P}_d=\left[ \boldsymbol{\xi}^{\left(d\right)}_{i_d},\boldsymbol{\eta}^{\left(d\right)}_{i_d} \right] \\
& \mathbf{g'}^{\left(d\right)}_{i_d}\triangleq -\overline{\mathbf{c'}}^{\left(d\right)}_{i_d}\overline{\mathbf{B'}}^{\left(d\right)^{\mathrm{T}}}_{i_d}\mathbf{P'}_d=\left[ \boldsymbol{\xi'}^{\left(d\right)}_{i_d},\boldsymbol{\eta'}^{\left(d\right)}_{i_d} \right]
\end{aligned}
\right.
\end{equation*}
such that $\boldsymbol{\xi}^{\left(d\right)}_{i_d},\; \boldsymbol{\xi'}^{\left(d\right)}_{i_d}\in \mathbb{R}^{\Gamma_d\Gamma_{d+1}}$, $\boldsymbol{\eta}^{\left(d\right)}_{i_d}\in \mathbb{R}^{R_dR_{d+1}-\Gamma_d\Gamma_{d+1}}$ and $\boldsymbol{\eta'}^{\left(d\right)}_{i_d}\in \mathbb{R}^{R'_d{R'}_{d+1}-\Gamma_d\Gamma_{d+1}}$.

We can deduce the solution (see Appendix A for details) as follows:
\begin{align}
& \left[\boldsymbol{\alpha}^{\left(d\right)}_{i_d},\boldsymbol{\beta}^{\left(d\right)}_{i_d},\boldsymbol{\gamma}^{\left(d\right)}_{i_d}\right]^* \notag \\
=& \mathop{\arg\min}_{\substack{\boldsymbol{\alpha}^{\left(d\right)}_{i_d},\boldsymbol{\beta}^{\left(d\right)}_{i_d}\\ \boldsymbol{\gamma}^{\left(d\right)}_{i_d}}} \frac{1}{2}\left[\boldsymbol{\alpha}^{\left(d\right)}_{i_d},\boldsymbol{\beta}^{\left(d\right)}_{i_d},\boldsymbol{\gamma}^{\left(d\right)}_{i_d}\right]\widehat{\mathbf{H}}^{\left(d\right)}_{i_d}\left[\boldsymbol{\alpha}^{\left(d\right)}_{i_d},\boldsymbol{\beta}^{\left(d\right)}_{i_d},\boldsymbol{\gamma}^{\left(d\right)}_{i_d}\right]^{\mathrm{T}} \notag \\
& +\left[\boldsymbol{\alpha}^{\left(d\right)}_{i_d},\boldsymbol{\beta}^{\left(d\right)}_{i_d},\boldsymbol{\gamma}^{\left(d\right)}_{i_d}\right]\left[\boldsymbol{\xi}^{\left(d\right)}_{i_d}+\boldsymbol{\xi'}^{\left(d\right)}_{i_d},\boldsymbol{\eta}^{\left(d\right)}_{i_d},\boldsymbol{\eta'}^{\left(d\right)}_{i_d}\right]^{\mathrm{T}}+ \notag \\
& \frac{1}{2}z^{\left(d\right)}_{i_d}+\frac{1}{2}{z'}^{\left(d\right)}_{i_d} \notag \\
=& -\widehat{\mathbf{g}}^{\left(d\right)}_{i_d}\widehat{\mathbf{H}}_{i_d}^{\left(d\right)^{\dagger}},
\label{update3}
\end{align}
where $\widehat{\mathbf{g}}^{\left(d\right)}_{i_d}=\left[ \boldsymbol{\xi}^{\left(d\right)}_{i_d}+\boldsymbol{\xi'}^{\left(d\right)}_{i_d},\boldsymbol{\eta}^{\left(d\right)}_{i_d},\boldsymbol{\eta'}^{\left(d\right)}_{i_d} \right]$ and the Hessian matrix is
\begin{align*}
& \widehat{\mathbf{H}}^{\left(d\right)}_{i_d}= \\
& \begin{bmatrix}
\widetilde{\mathbf{H}}_{i_d}^{\left(d\right)11}+\widetilde{\mathbf{H'}}_{i_d}^{\left(d\right)11} & \widetilde{\mathbf{H}}_{i_d}^{\left(d\right)12}+\widetilde{\mathbf{H}}_{i_d}^{\left(d\right)21^{\mathrm{T}}} & \widetilde{\mathbf{H'}}_{i_d}^{\left(d\right)12}+\widetilde{\mathbf{H'}}_{i_d}^{\left(d\right)21^{\mathrm{T}}} \\
\widetilde{\mathbf{H}}_{i_d}^{\left(d\right)21}+\widetilde{\mathbf{H}}_{i_d}^{\left(d\right)12^{\mathrm{T}}} & \widetilde{\mathbf{H}}_{i_d}^{\left(d\right)22} & \mathbf{0} \\
\widetilde{\mathbf{H'}}_{i_d}^{\left(d\right)21}+\widetilde{\mathbf{H'}}_{i_d}^{\left(d\right)12^{\mathrm{T}}} & \mathbf{0} & \widetilde{\mathbf{H'}}_{i_d}^{\left(d\right)22}
\end{bmatrix}
.
\end{align*}

The details for this block coordinate descent method for coupled TR completion is outlined in Algorithm \ref{algorithm_CTRC}.
\begin{algorithm}
\caption{BCD for CTRC}
\label{algorithm_CTRC}
\begin{algorithmic}[1]
\REQUIRE Two ground truth tensors $\mathcal{T}_1$ and $\mathcal{T}_2$, two binary tensors $\mathcal{W}_1$ and $\mathcal{W}_2$, the maximal iteration $K$
\ENSURE Two recovered tensors $\mathcal{X}$ and $\mathcal{Y}$, two sets of TR factors $\left\{\mathcal{U}\right\}$ and $\left\{\mathcal{V}\right\}$
\STATE Apply Algorithm 1 to initialize $\left\{\mathcal{U}\right\}$ and $\left\{\mathcal{V}\right\}$ 
\FOR{$k=1$ \TO $K$}
\STATE Update the uncoupled TR factors of $\mathfrak{R}_1$ according to (\ref{update1})
\STATE Update the uncoupled TR factors of $\mathfrak{R}_2$ according to (\ref{update2})
\STATE Update the coupled TR factors of $\mathfrak{R}_1$ and $\mathfrak{R}_2$ according to (\ref{update3})
\STATE Update $\mathcal{X}=\mathfrak{R}\left(\left\{\mathcal{U}\right\}\right)$, $\mathcal{Y}=\mathfrak{R}\left(\left\{\mathcal{V}\right\}\right)$
\IF{converged}
\STATE break
\ENDIF
\ENDFOR
\RETURN{$\mathcal{X}$, $\mathcal{Y}$, $\left\{\mathcal{U}\right\}$, $\left\{\mathcal{V}\right\}$}
\end{algorithmic}
\end{algorithm}

Note that this algorithm can be easily extended to the case where more than two tensor rings are coupled, and only the scheme for updating the coupled components is changed in the generalized cases. We have the Hessian matrix defined in the form of block matrix as follows:
\begin{equation*}
\left\{
\begin{aligned}
& \widehat{\mathbf{H}}^{\left(d\right)}_{i_d}\left\{1,1\right\}=\sum^N_{n=1}\widetilde{\mathbf{H}^{\left(n\right)}}_{i_d}^{\left(d\right)11},\; \widehat{\mathbf{H}}_{i_d}\left\{n,n\right\}=\widetilde{\mathbf{H}^{\left(n\right)}}_{i_d}^{\left(d\right)22} \\
& \left(n=2,\dotsc,N\right), \\
& \widehat{\mathbf{H}}^{\left(d\right)}_{i_d}\left\{1,n\right\}=\widehat{\mathbf{H}}_{i_d}\left\{n,1\right\}^{\mathrm{T}}=\widetilde{\mathbf{H}^{\left(n\right)}}_{i_d}^{\left(d\right)12}+\widetilde{\mathbf{H}^{\left(n\right)}}_{i_d}^{\left(d\right)21^{\mathrm{T}}} \\
& \left(n=2,\dotsc,N\right), \\
& \widehat{\mathbf{H}}^{\left(d\right)}_{i_d}\left\{m,n\right\}=\mathbf{0}\; \left(m\neq n,\;m\neq 1,\;n\neq 1\right)
\end{aligned}
\right.
\end{equation*}
and $\widehat{\mathbf{g}}^{\left(d\right)}_{i_d}=\left[ \sum^N_{n=1}\boldsymbol{\xi^{\left(n\right)}}^{\left(d\right)}_{i_d},\boldsymbol{\eta}^{\left(d\right)}_{i_d},\boldsymbol{\eta^{\left(N\right)}}^{\left(d\right)}_{i_d} \right]$, where $\widetilde{\mathbf{H}^{\left(n\right)}}^{\left(d\right)}_{i_d}=\mathbf{P}^{{\left(n\right)}^{\mathrm{T}}}_d\mathbf{H}^{\left(n\right)}_{i_d}\mathbf{P}^{\left(n\right)}_d$ and
\begin{align*}
\widetilde{\mathbf{g}^{\left(n\right)}}^{\left(d\right)}_{i_d}=-\widetilde{\mathbf{c}^{\left(n\right)}}^{\left(d\right)}_{i_d}\mathbf{B}_d^{{\left(n\right)}^{\mathrm{T}}}\mathbf{P}^{\left(n\right)}_d=\left[ \boldsymbol{\xi^{\left(n\right)}}^{\left(d\right)}_{i_d},\boldsymbol{\eta^{\left(n\right)}}^{\left(d\right)}_{i_d} \right].
\end{align*}

\subsection{Algorithmic Complexity}

Assuming all the tensors $\mathcal{X}_1$, $\dotsc$, $\mathcal{X}_N$ have the same size $I_1\times \dotsc \times I_D$ and TR rank $\left[R,\dotsc,R\right]$. Supposing the number of the samples for the $n$-th tensor is $m$. The computation of Hessian matrix $\mathbf{H}^{\left(d\right)}_{i_d}$ costs $O\left(\mathrm{SR}\times R^4\prod^D_{k=1,\; k\neq d}I_d\right)=O\left(mR^4/I_d\right)$, where $\mathrm{SR}$ is the sampling rate for $\mathcal{X}_n$. Thus updating the $d$-th TR factor costs $O\left(mR^4\right)$ and one iteration  costs $O\left(mNDR^4\right)$.

The computation of $\mathbf{H}^{\left(d\right)^\dagger}_{i_d}$ costs $O\left(R^6\right)$ and the update of the $d$-th TR factor costs $O\left(I_dR^6\right)$. Hence one iteration  costs $O\left(NR^6\sum^D_{d=1}I_d\right)$.

The total computational cost of one iteration of BCD for CTRC is $\max\left\{O\left(mNDR^4\right),O\left(NR^6\sum^D_{d=1}I_d\right)\right\}=O\left(mNDR^4\right)$. The storage cost is $O\left(NDR^2\right)$.

\subsection{Convergence Analysis}

Objective function (\ref{model-coupled tensor ring completion}) is real analytic since it is essentially a polynomial function, hence it is easy to verify that (\ref{model-coupled tensor ring completion}) satisfy the Kurdyka-Lojasiewicz inequality with $\theta=1/2$ \cite{xu2013block}. As a consequence, the asymptotic convergence rate of Algorithm \ref{algorithm_CTRC} is linear.

\section{Excess Risk Bound}
\label{section_bound}

We define $\bar{l}_{\mathbb{T}}\left(\cdot,\cdot\right)$ as the average of the perfect square trinomial $l\left(\cdot,\cdot\right)$ computed on a finite training set $\mathbb{T}$. For concise expression of average test error, we use notation $\bar{l}_{\mathbb{T}}\left( \left\{\mathcal{X},\mathcal{Y}\right\},\left\{\mathcal{T}_1,\mathcal{T}_2\right\} \right)$ to denote the average training error over $\mathbb{T}$, where we refer to $\mathbb{T}\subseteq \mathbb{O}$ as the union of $\mathbb{T}_1\subseteq \mathbb{O}_1$ and $\mathbb{T}_1\subseteq \mathbb{O}_2$. Similarly, we can define $\bar{l}_{\mathbb{S}}\left(\mathcal{X},\mathcal{Y}\right)$ as the average test error measured by $l\left(\cdot,\cdot\right)$ over $\mathbb{S}\subseteq \mathbb{O}^\perp$. As in \cite{tolstikhin2015permutational}, we assume that $\left| \mathbb{S}_i \right|=\left| \mathbb{T}_i \right|$ for any $i\in \left\{1,2\right\}$.

We use the assumption that each TR factor $X$ obeys an i.i.d. conditional normal distribution with mean $0$ and precision $T^{-1}$ \cite{zhao2015bayesian}, where $T$ is governed by a gamma distribution with parameters $a$ and $b$. Multiplying two probability density functions (PDFs), the joint distribution turns out an i.i.d. normal-gamma distribution. The marginal distribution of $X$ is an i.i.d. non-standardized Student's t-distribution
\begin{align}
p\left(x;a,b\right)=\frac{\Gamma\left(a+\frac{1}{2}\right)}{\sqrt{2\pi b}\Gamma\left(a\right)}\left(1+\frac{x^2}{2b}\right)^{-\left(a+\frac{1}{2}\right)},
\end{align}
where $\Gamma\left(\cdot\right)$ is the gamma function.

Given an assumption that $\mathcal{X}=\mathfrak{R}\left(\left\{\mathcal{U}\right\}\right)$ with TR rank $\left[R,\dotsc,R\right]$ and each TR factor is an independent Student's t random tensor, we can define a hypothesis class $\mathcal{H}\triangleq \left\{\mathcal{X},\mathcal{Y}\;|\;\mathcal{U}^{\left(d_1\right)}\sim \mathcal{S}\left(a,b\right),\mathcal{V}^{\left(d_2\right)}\sim \mathcal{S}\left(a,b\right) \right\}$, where the symbol $\mathcal{S}\left(a,b\right)$ represents non-standardized Student's t-distribution with with parameters $a$ and $b$.

Without loss of generality, we assume $l\left(\cdot,\cdot\right)$ is $\Lambda$-Lipschitz continuous since the F-norms of two tensors are centralized with overwhelming probability. By leveraging the recently proposed permutational Rademacher complexity \cite{tolstikhin2015permutational}, the following theorem characterizes the excess risk of coupled TR completion.
\begin{theorem}
Under the hypothesis $\mathcal{H}$ mentioned before, the excess risk of the coupled TR completion (\ref{model-coupled tensor ring completion}) is bounded as
\begin{align}
& \bar{l}_{\mathbb{S}}\left( \left\{\mathcal{X},\mathcal{Y}\right\},\left\{\mathcal{T}_1,\mathcal{T}_2\right\} \right)-\bar{l}_{\mathbb{T}}\left( \left\{\mathcal{X},\mathcal{Y}\right\},\left\{\mathcal{T}_1,\mathcal{T}_2\right\} \right)\leq \notag \\
& \Lambda\left(1+\frac{2}{\sqrt{2\pi \left| \mathbb{T} \right|}-2}\right) \left(\frac{df^*_2b}{df^*_1a}\right)^{\frac{D_2}{2}} \frac{\operatorname{B}^{L}\left(df^*_1+\frac{1}{2},df^*_2-\frac{1}{2}\right)}{\sqrt{\left|\mathbb{T}_2\right|} \operatorname{B}^{L}\left(df^*_1,df^*_2\right)} \notag \\
& {_{\substack{D_1+D_2 \\ -2L+1}}}F_{\substack{D_1+D_2 \\ -2L}} \left( \substack{df^*_1,\dotsc,df^*_2-\frac{1}{2},\dotsc,-\frac{1}{2} \\ 1-df^*_2,\dotsc,\frac{1}{2}-df^*_1,\dotsc} \Bigg| -\frac{\left|\mathbb{T}_2\right|}{\left|\mathbb{T}_1\right|}\left(-\frac{df^*_2b}{df^*_1a}\right)^{D_1-D_2} \right) \notag \\
& + \sqrt{\frac{2\left| \mathbb{T}\cup \mathbb{S} \right| \ln\left(\frac{1}{\delta}\right)}{\left(\left| \mathbb{T}\cup \mathbb{S} \right|-\frac{1}{2}\right)^2}}
\label{excess risk-coupled completion1}
\end{align}
and
\begin{align}
& \bar{l}_{\mathbb{S}}\left( \left\{\mathcal{X},\mathcal{Y}\right\},\left\{\mathcal{T}_1,\mathcal{T}_2\right\} \right)-\bar{l}_{\mathbb{T}}\left( \left\{\mathcal{X},\mathcal{Y}\right\},\left\{\mathcal{T}_1,\mathcal{T}_2\right\} \right)\leq \notag \\
& \Lambda\left(1+\frac{2}{\sqrt{2\pi \left| \mathbb{T} \right|}-2}\right) \left(\frac{df^*_2b}{df^*_1a}\right)^{\frac{D_1}{2}} \frac{\operatorname{B}^{L}\left(df^*_1+\frac{1}{2},df^*_2-\frac{1}{2}\right)}{\sqrt{\left|\mathbb{T}_1\right|} \operatorname{B}^{L}\left(df^*_1,df^*_2\right)} \notag \\
& {_{\substack{D_1+D_2 \\ -2L+1}}}F_{\substack{D_1+D_2 \\ -2L}} \left( \substack{df^*_2-\frac{1}{2},\dotsc,df^*_1,\dotsc,-\frac{1}{2} \\ \frac{1}{2}-df^*_1,\dotsc,1-df^*_2,\dotsc} \Bigg| -\frac{\left|\mathbb{T}_1\right|}{\left|\mathbb{T}_2\right|}\left(-\frac{df^*_2b}{df^*_1a}\right)^{D_2-D_1} \right) \notag \\
& \sqrt{\frac{2\left| \mathbb{T}\cup \mathbb{S} \right| \ln\left(\frac{1}{\delta}\right)}{\left(\left| \mathbb{T}\cup \mathbb{S} \right|-\frac{1}{2}\right)^2}}
\label{excess risk-coupled completion2}
\end{align}
alternatively with probability at least $1-\delta$. Here the definitions of two parameters $df^*_1$ and $df^*_2$ can refer to (\ref{equation-3_1}) and (\ref{equation-3_2}) in Appendix A. The usage of (\ref{excess risk-coupled completion1}) or (\ref{excess risk-coupled completion2}) depends on the input value of the generalized hypergeometric function (for detail please refer to Appendix B). Moreover, with the same probability the excess risk of each individual TR completion is bounded by
\begin{align}
& \Lambda\left(1+\frac{2}{\sqrt{2\pi \left| \mathbb{T}_n \right|}-2}\right) \left(\frac{df^*_2b}{df^*_1a}\right)^{\frac{D_n}{2}} \frac{\operatorname{B}^{D_n}\left(df^*_1+\frac{1}{2},df^*_2-\frac{1}{2}\right)}{\sqrt{\left|\mathbb{T}_n\right|} \operatorname{B}^{D_n}\left(df^*_1,df^*_2\right)} \notag \\
& + \sqrt{\frac{2\left| \mathbb{T}_n\cup \mathbb{S}_n \right| \ln\left(\frac{1}{\delta}\right)}{\left(\left| \mathbb{T}_n\cup \mathbb{S}_n \right|-\frac{1}{2}\right)^2}},\; n=1,2.
\label{excess risk-individual completion}
\end{align}
\label{theorem_risk bound}
\end{theorem}

Theorem \ref{theorem_risk bound} reports a phenomenon that the risk bound of coupled completion can be much lower than that of individual completion. It suffices to illustrate this by comparing their bounds term by term since these expressions are multiplicative. Note that the $F$ function approaches to $1$ if the absolute value of input argument is close to $0$. Supposing $D_1\geq D_2$ without loss of generality, the $F$ function can yield a number close to $1$ by choosing $\left| \mathbb{T}_1 \right|$ and $\left| \mathbb{T}_2 \right|$ such that $\left| \mathbb{T}_1 \right|$ is larger enough than $\left| \mathbb{T}_2 \right|$. Thus the risk bound (\ref{excess risk-coupled completion1}) or (\ref{excess risk-coupled completion2}) can be much lower than the maximum of the bounds (\ref{excess risk-individual completion}).

To discuss the effect of these parameters on the risk bound, note that $\mathbb{E}\left[\sqrt{X}\right]\leq \sqrt{\mathbb{E}\left[X\right]}$, we resort to a supremum of the risk bound that stems from Appendix B:
\begin{align}
& \Lambda\left(1+\frac{2}{\sqrt{2\pi \left| \mathbb{T} \right|}-2}\right) \left(\frac{df^*_2b}{df^*_1a}\right)^{\frac{L}{2}} \frac{\operatorname{B}^{L}\left(df^*_1+\frac{1}{2},df^*_2-\frac{1}{2}\right)}{\operatorname{B}^{L}\left(df^*_1,df^*_2\right)} \notag \\
& \sqrt{\frac{\left(\frac{kb}{a-1}\right)^{D_1-L}}{\left| \mathbb{T}_1 \right|}+\frac{\left(\frac{kb}{a-1}\right)^{D_2-L}}{\left| \mathbb{T}_2 \right|}} + \sqrt{\frac{2\left| \mathbb{T}_n\cup \mathbb{S}_n \right| \ln\left(\frac{1}{\delta}\right)}{\left(\left| \mathbb{T}\cup \mathbb{S} \right|-\frac{1}{2}\right)^2}},
\label{supremum of risk bound}
\end{align}
where $k=IR^2$. Therefore, increasing $\left| \mathbb{T}_1 \right|$ or $\left| \mathbb{T}_2 \right|$ is beneficial to the recovery performance, but the increments of $I$, $R$, $D_1$ and $D_2$ are unfavorable. To examine the effect of $L$, we reformulate the above term that contains $L$ in (\ref{supremum of risk bound}) as
\begin{align*}
\left[ \frac{\sqrt{\frac{df^*_2b}{df^*_1a}} \frac{\operatorname{B}\left(df^*_1+\frac{1}{2},df^*_2-\frac{1}{2}\right)}{\operatorname{B}\left(df^*_1,df^*_2\right)}}{\sqrt{\frac{kb}{a-1}}} \right]^{L} \sqrt{\frac{\left(\frac{kb}{a-1}\right)^{D_1}}{\left| \mathbb{T}_1 \right|}+\frac{\left(\frac{kb}{a-1}\right)^{D_2}}{\left| \mathbb{T}_2 \right|}},
\end{align*}
where the left term is recognized as an exponential function with base $\mathbb{E}\left[\sqrt{X}\right]/\sqrt{\mathbb{E}\left[X\right]}$, hence increasing the number of coupled dimensions improves the recovery performance.

This result of coupled completion can be also regarded from the viewpoint of mutual information. 
$I\left(\mathcal{X};\mathcal{Y}\right)=\iint p\left(\mathcal{X},\mathcal{Y}\right)\ln \frac{p\left(\mathcal{X},\mathcal{Y}\right)}{p\left(\mathcal{X}\right)p\left(\mathcal{Y}\right)} d\mathcal{X}d\mathcal{Y}$.
In the scenario that two TRs are not coupled, they have zero mutual information, thus they cannot help each other's recovery. On the other hand, it becomes the differential entropy if they are totally coupled. The amount of information reaches the maximum, which results in the lowest recovery deviation. The summation of the Hessian matrices in Algorithm \ref{algorithm_CTRC} reflects the joint entropy, where the amplitude ratio of the Hessian matrices is $\left| \mathbb{T}_1 \right| : \left| \mathbb{T}_2 \right| : \dotsm : \left| \mathbb{T}_N \right|$.

By leveraging the Gautschi's inequality, we have the following bound:
\begin{align*}
\frac{\operatorname{B}\left(df^*_1+\frac{1}{2},df^*_2-\frac{1}{2}\right)}{\operatorname{B}\left(df^*_1,df^*_2\right)} \leq \sqrt{\frac{df^*_1+\frac{1}{2}}{df^*_2-1}}.
\end{align*}
Consequently, the excess risk of our coupled F-norm is bound by $O\left( \left( df^*_2 / df^*_1 \right)^{\left(D-L\right)/2} \right)$, which is on the order of $O\left(R^{D-L}I^{\left(D-L\right)/2}\right)$. As a comparison, the bound in \cite{wimalawarne2018efficient} has the order $O\left(D16^DI^{\left(D+1\right)/2}R\ln^{D-1/2}\left(I\right)\right)$ by considering bounding the nuclear norm. We notice that the optimal ranks $R$ of TR decomposition and TK decomposition for a tensor are usually different, which can be theoretically proved by \cite{zniyed2020high} and empirically corroborated by the evidence in \cite{huang2019low}.

\section{Numerical Experiments}
\label{section_experiment}

In this section, the proposed algorithm is evaluated on two kinds of datasets, i.e., synthetic data and real-world data. The synthetic dataset is employed to verify the theoretical results, and real-world data based experiments are used to test the empirical performance of the proposed CTRC and three other state-of-the-art algorithms, including the coupled nuclear norm minimization for coupled tensor completion (CNN) \cite{wimalawarne2018efficient}, advanced coupled matrix and tensor factorization (ACMTF) \cite{acar2013structure, acar2014structure} structured data fusion by nonlinear least squares (SDF) \cite{sorber2015structured}. The low rank tensor completion via alternating least square (TR-ALS) \cite{wang2017efficient} is also compared as a baseline, since it can be regarded as the individual TR completion.

The root of mean square error (RMSE) defined as $\text{RMSE}=\lVert \hat{\mathcal{X}}-\mathcal{X} \rVert_{\mathrm{F}}/{\sqrt{\left| \mathcal{X} \right|}}$ is used to measure the completion accuracy, where $\mathcal{X}$ is the ground-truth and $\hat{\mathcal{X}}$ is the estimate of $\mathcal{X}$. We use computational CPU time (in seconds) as a measure of algorithmic complexity.

The sampling rate (SR) is defined as the ratio of the number of samples to the total number of the elements of tensor $\mathcal{X}$, which is denoted as $\text{SR}=\left|\mathbb{O}\right|/\left|\mathcal{X}\right|$. For fair comparison, the parameters in each algorithm are tuned to give optimal performance. For the proposed BCD for CTRC, one of the stop criteria is that the relative change $\text{RC}=\lVert \mathcal{X}_k-\mathcal{X}_{k-1}\rVert_{\mathrm{F}}/\lVert \mathcal{X}_{k-1}\rVert_{\mathrm{F}}$ is less than a tolerance that is set to $1\times 10^{-8}$. We set the maximal number of iterations $K=200$ in experiments on synthetic data and $K=100$ in experiments on real-world data.

All the experiments are conducted in MATLAB 9.7.0 on a computer with a 2.8GHz CPU of Intel Core i7 and a 16GB RAM.

\subsection{Synthetic Data}

In this section, we test our algorithm on randomly generated tensor data for completion problem. We generate two tensors of the same size $20\times 20\times 20\times 20$ using the TR decomposition (\ref{TR decomposition}). The TR factors are randomly sampled from the standard normal distribution, i.e., $\mathcal{U}^{\left(d\right)}\left(r_d,i_d,r_{d+1}\right)\sim \mathcal{N}\left(0,1\right)$, $\mathcal{V}^{\left(d\right)}\left(r_d,i_d,r_{d+1}\right)\sim \mathcal{N}\left(0,1\right)$, $d=1,\dotsc,4$. Then we couple two tensor rings by setting $\mathcal{U}^{\left(d\right)}=\mathcal{V}^{\left(d\right)}$, $d=1,\dotsc,3$. We compute the tensors $\mathcal{T}_1$ and $\mathcal{T}_2$ according to these factors. 

The proposed algorithm's performance is evaluated by the phase transition on TR rank versus sampling rate of tensor $\mathcal{T}_1$ under different settings of sampling rate of tensor $\mathcal{T}_2$ and the number of coupled TR factors. The sampling rate of $\mathcal{T}_1$ ranges from $0.005$ to $0.1$ with interval $0.005$, and the sampling rate of $\mathcal{T}_2$ ranges from $0.05$ to $0.2$ with interval $0.05$. The TR rank varies from $2$ to $8$, and the number of coupled TR factors is $1$, $2$ and $3$.

\begin{figure*}[htbp]
\centering
\begin{subfigure}[t]{0.45\textwidth}
\centering
\setlength{\abovecaptionskip}{0pt}
\setlength{\belowcaptionskip}{-2pt}
\includegraphics[scale=0.2]{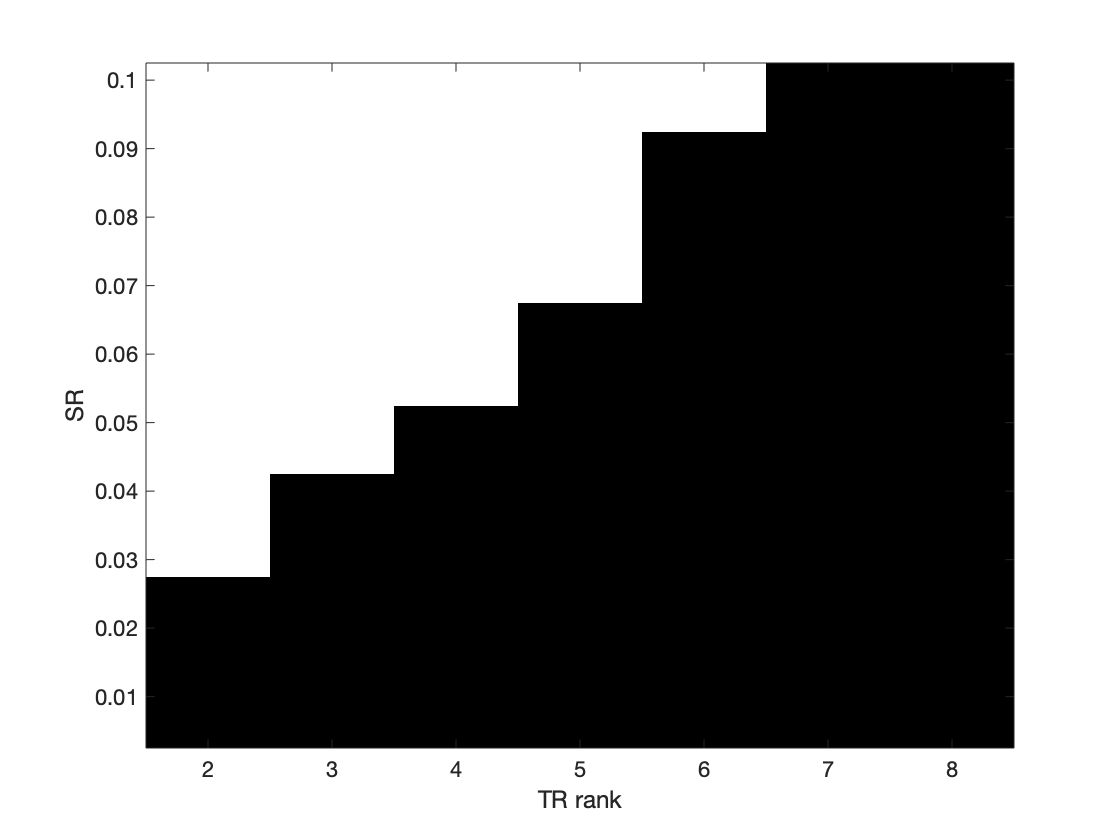}
\subcaption{The phase transition of an individual completion.}
\end{subfigure}
\qquad\qquad
\begin{subfigure}[t]{0.45\textwidth}
\centering
\setlength{\abovecaptionskip}{0pt}
\setlength{\belowcaptionskip}{-2pt}
\includegraphics[scale=0.2]{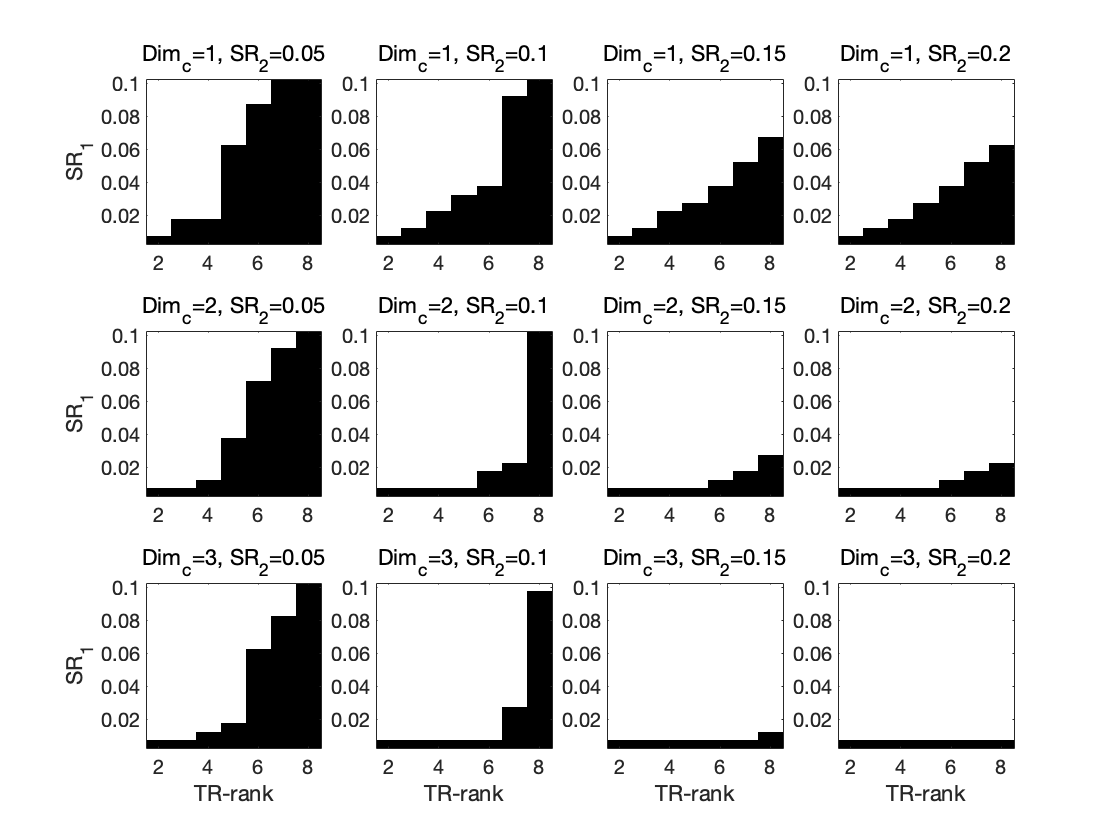}
\subcaption{The phase transition on TR rank versus sampling rate of tensor $\mathcal{T}_1$ with various sampling rates of tensor $\mathcal{T}_2$ and the numbers of coupled TR factors.}
\end{subfigure}
\caption{The exact recovery result of randomly generated data with random sampling.}
\label{result-synthetic1}
\end{figure*}

The results are shown in Fig. \ref{result-synthetic1}, where $\text{Dim}_{\text{c}}$  represents the number of the coupled TR factors, and $\text{SR}_1$ and $\text{SR}_2$ represent the sampling rates of $\mathcal{T}_1$ and $\mathcal{T}_2$, respectively. In phase transition, the white patch means a successful recovery whose RMSE is less than $1\times 10^{-6}$, and the black patch means a failure. The successful area increases when sampling rate of $\mathcal{T}_2$ or the number of the coupled TR factors increases. This is because the first TR can benefit from the second one with increasing sampling rate or number of coupled factors, even though the number of samples of $\mathcal{T}_1$ is beyond its sampling limit in the individual case. Numerically, the ratio of the magnitude of Hessian matrices $\widetilde{\mathbf{H}}_{i_d}$ and $\widetilde{\mathbf{H'}}_{i_d}$ generated from two tensors is $\text{SR}_1/\text{SR}_2$. This suggests the tensor with larger number of samples would be dominated in the updating scheme, which is the reason for the breaking of the sampling limit.

\subsection{The UCLAF Data}

In this subsection, the user-centered collaborative location and activity filtering (UCLAF) data \cite{zheng2010collaborative} is used. It comprises $164$ users' GPS trajectories based on their partial $168$ locations and $5$ activity annotations. In the experiment, we only use the link activity, i.e., only entry greater than $0$ is set to $1$. users with no data are discarded, which results in a new tensor of size $144\times 168\times 5$. The user-location matrix which has size $144\times 168$ is coupled with this tensor as a side information. We randomly choose $50\%$ samples from the tensor and the matrix independently. For each algorithm we conduct $10$ experiments for avoiding fortuitous result. We set the first dimension as the coupled dimension.
\begin{figure*}[htbp]
\centering
\begin{subfigure}[t]{0.45\textwidth}
\centering
\setlength{\abovecaptionskip}{0pt}
\setlength{\belowcaptionskip}{-2pt}
\includegraphics[scale=0.22]{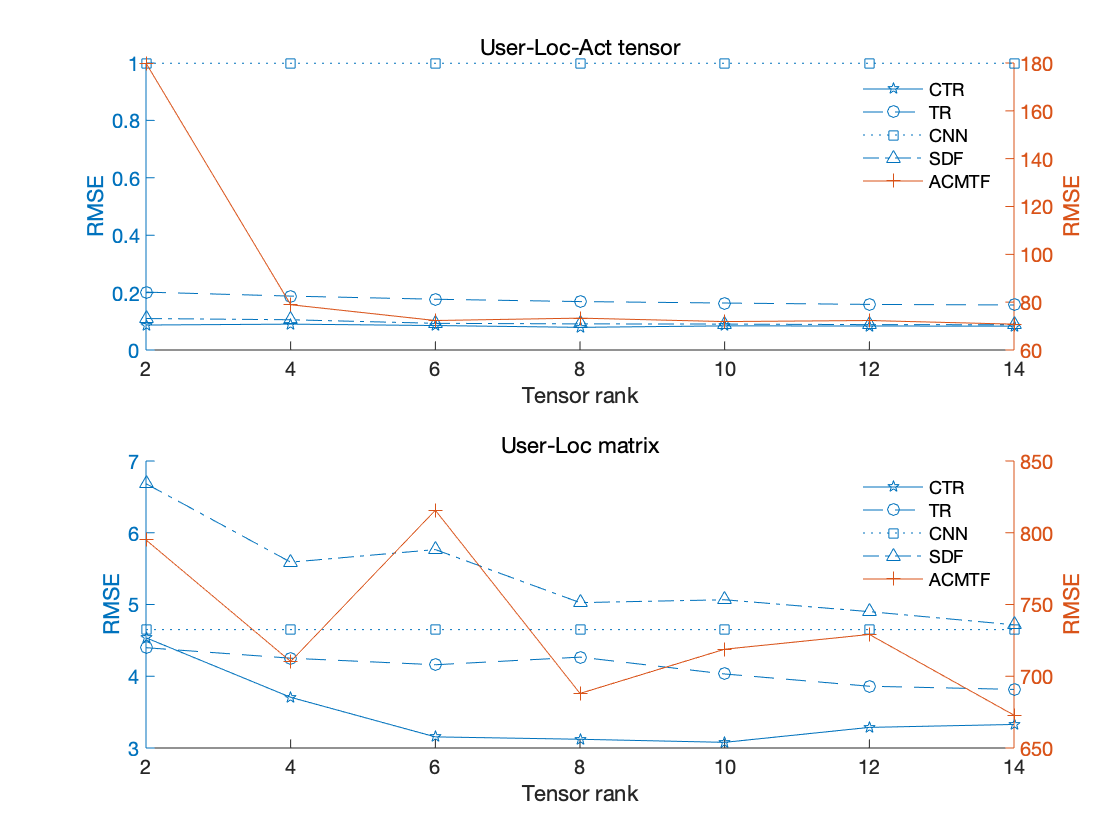}
\subcaption{RMSE}
\end{subfigure}
\qquad\qquad
\begin{subfigure}[t]{0.45\textwidth}
\centering
\setlength{\abovecaptionskip}{0pt}
\setlength{\belowcaptionskip}{-2pt}
\includegraphics[scale=0.22]{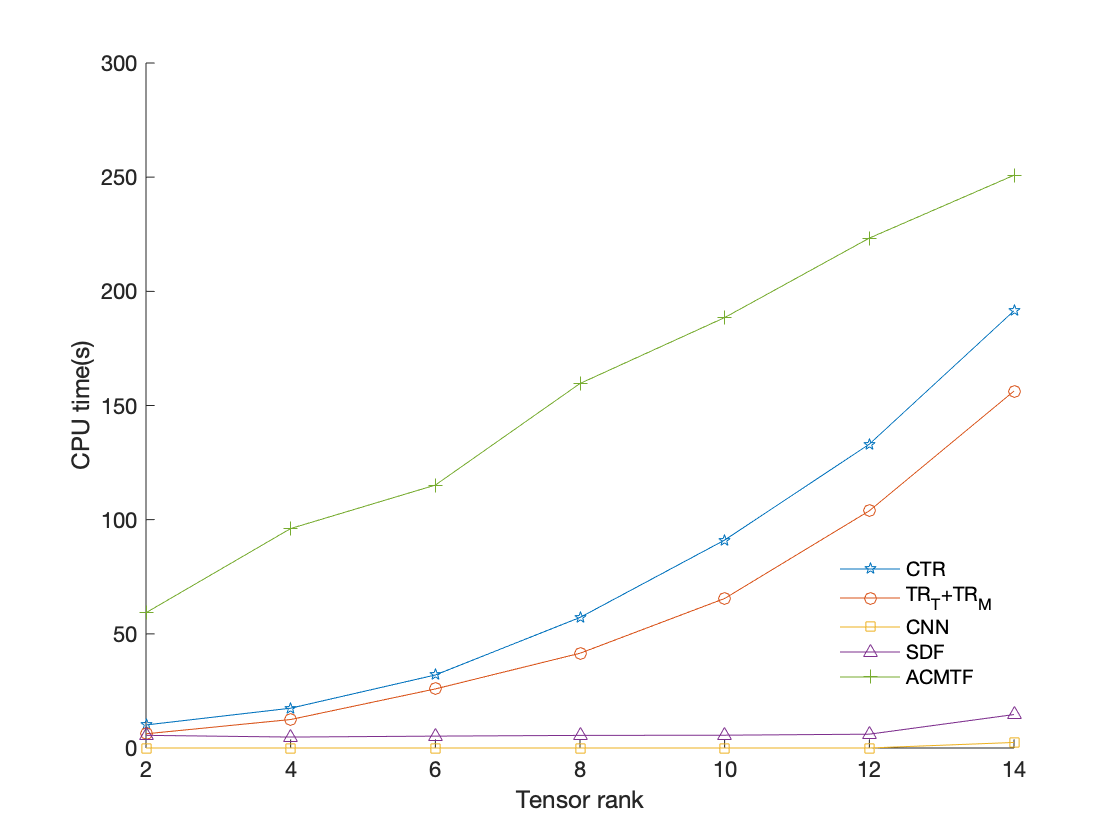}
\subcaption{CPU time (s)}
\end{subfigure}
\caption{The completion result of the UCLAF data derived by five algorithms. The panel (a) is the RMSE comparison and the panel (b) is the elapsed CPU time in seconds.}
\label{result-real-data1}
\end{figure*}

Fig. \ref{result-real-data1} shows the average completion accuracy result versus tensor rank  by five algorithms, where the accuracy is measured by RMSE. The label ``TR'' means the individual tensor ring completion method which is a baseline for comparing with the coupled tensor ring completion methods. From the result, the coupled completion method performs better than the individual one. The proposed CTRC shows lower RMSEs in both user-location-activity tensor completion and user-location matrix completion, which illustrates that the coupled TR's F-norm can lead to better performance compared with the other coupled norms.

\subsection{The SW-NIR Data}

A dataset consists of a set of short-wave near-infrared (SW-NIR) spectrums measured on an HP diode array spectrometer is used in this subsection \cite{wulfert1998influence}. It is used to trace the influence of the temperature on vibrational spectrum and the consequences for the predictive ability of multivariate calibration models. The spectrum of $19$ mixtures of ethanol, water, isopropanol and pure compounds are recorded in a $1$ cm cuvette at different temperatures, i.e., $30$, $40$, $50$, $60$ and $70$ degrees Celsius. We stack each spectrum recorded at different temperatures in the third dimension and forms a tensor of size $512\times 22\times 5$. The coupled matrix is derived by stacking the temperature records in a similar way, which results in a matrix of size $22\times 5$. We randomly choose $50\%$ entries from the tensor and the matrix independently. For each algorithm we repeatedly conduct $10$ experiments. We suppose the first dimension is coupled.
\begin{figure*}[htbp]
\centering
\begin{subfigure}[t]{0.45\textwidth}
\centering
\setlength{\abovecaptionskip}{0pt}
\setlength{\belowcaptionskip}{-2pt}
\includegraphics[scale=0.22]{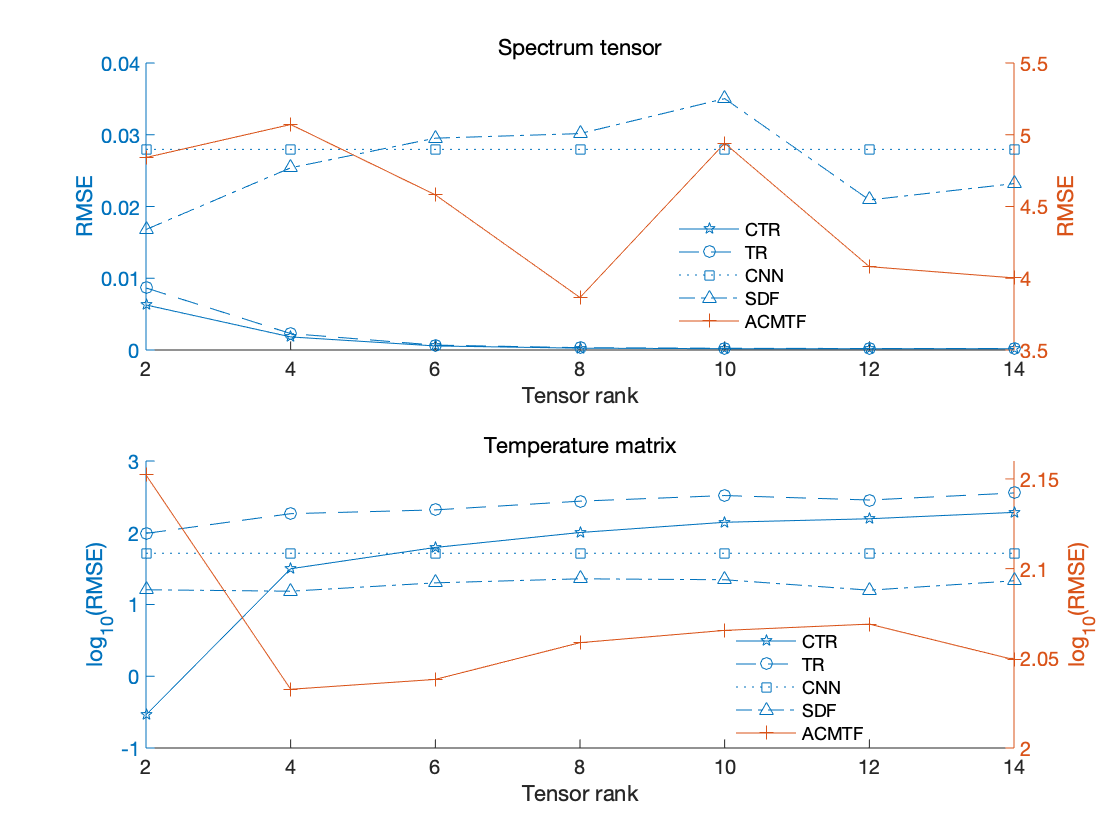}
\subcaption{RMSE}
\end{subfigure}
\qquad\qquad
\begin{subfigure}[t]{0.45\textwidth}
\centering
\setlength{\abovecaptionskip}{0pt}
\setlength{\belowcaptionskip}{-2pt}
\includegraphics[scale=0.22]{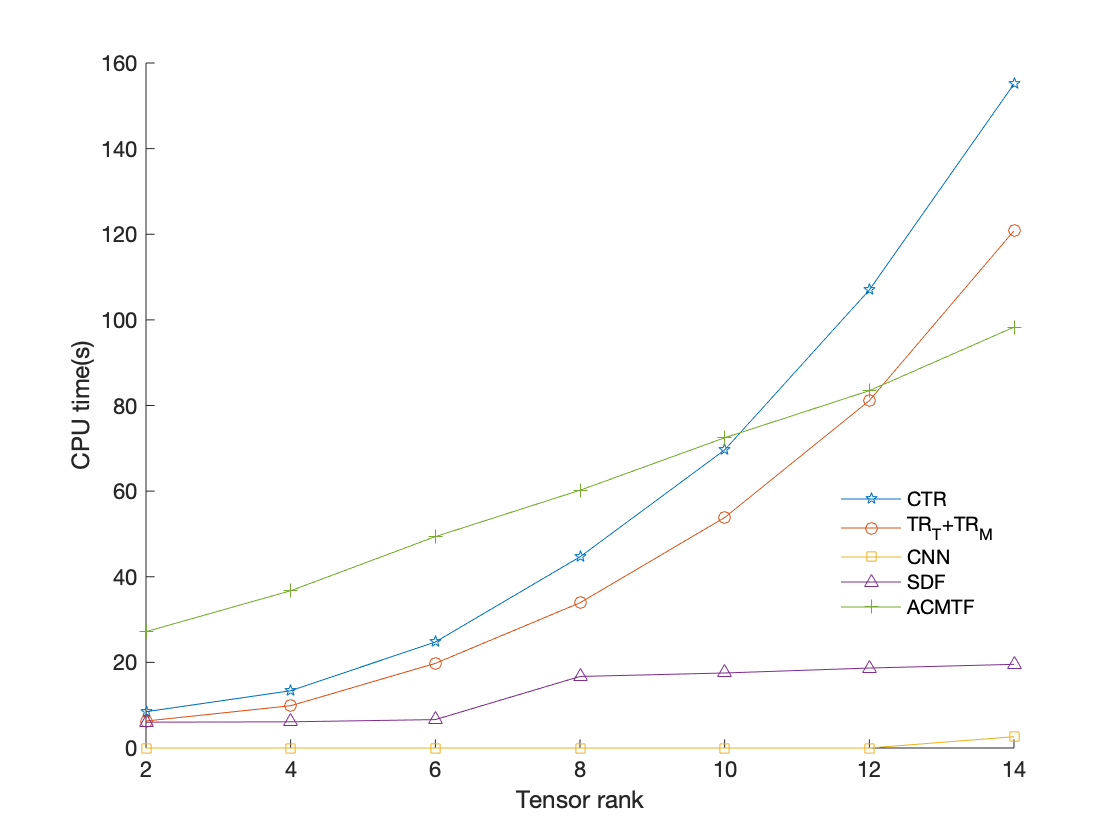}
\subcaption{CPU time (s)}
\end{subfigure}
\caption{The completion result of the SW-NIR data derived by five algorithms. The panel (a) is the RMSE comparison and the panel (b) is the elapsed CPU time in seconds.}
\label{result-real-data2}
\end{figure*}
 
Fig. \ref{result-real-data2} provides the completion results. It can be seen that the proposed CTRC generates the lowest RMSE for both the spectrum tensor and the temperature matrix when TR rank is $2$. The performance deteriorates when the TR rank becomes larger, which may imply the overfitting occurs \cite{wang2017efficient}.
The individual TR completion method performs worse than the coupled ones, which indicates the effectiveness of our method.

\subsection{The Licorice Data}

The dataset called ``Licorice'' used in this subsection \footnote{http://www.models.life.ku.dk/3Dnosedata} consists of $6$ good licorice samples, $6$ bad licorice samples and $6$ fabricated bad licorice samples. The time mode is a continuous time scale where the sensor signal has been measured every $0.5$ seconds from $0$ to $120$ seconds. The first time is the baseline signal (i.e. signal of carrier gas). Twelve sensors, all based on Metal Oxide Semiconductor (MOS) technologies, were used to register the volatile compounds from the samples. The data process is as follows. We select $6$ good and $6$ bad class samples and for each class we exclude the carrier gas signal, which forms two tensors of the same size $6 \times 240 \times 12$. We randomly select $50\%$ entries from each tensor, and for each algorithm we run $10$ times independently. In this experiment, we assume the last two dimensions are coupled.

\begin{figure*}[htbp]
\centering
\begin{subfigure}[t]{0.45\textwidth}
\centering
\setlength{\abovecaptionskip}{0pt}
\setlength{\belowcaptionskip}{-2pt}
\includegraphics[scale=0.22]{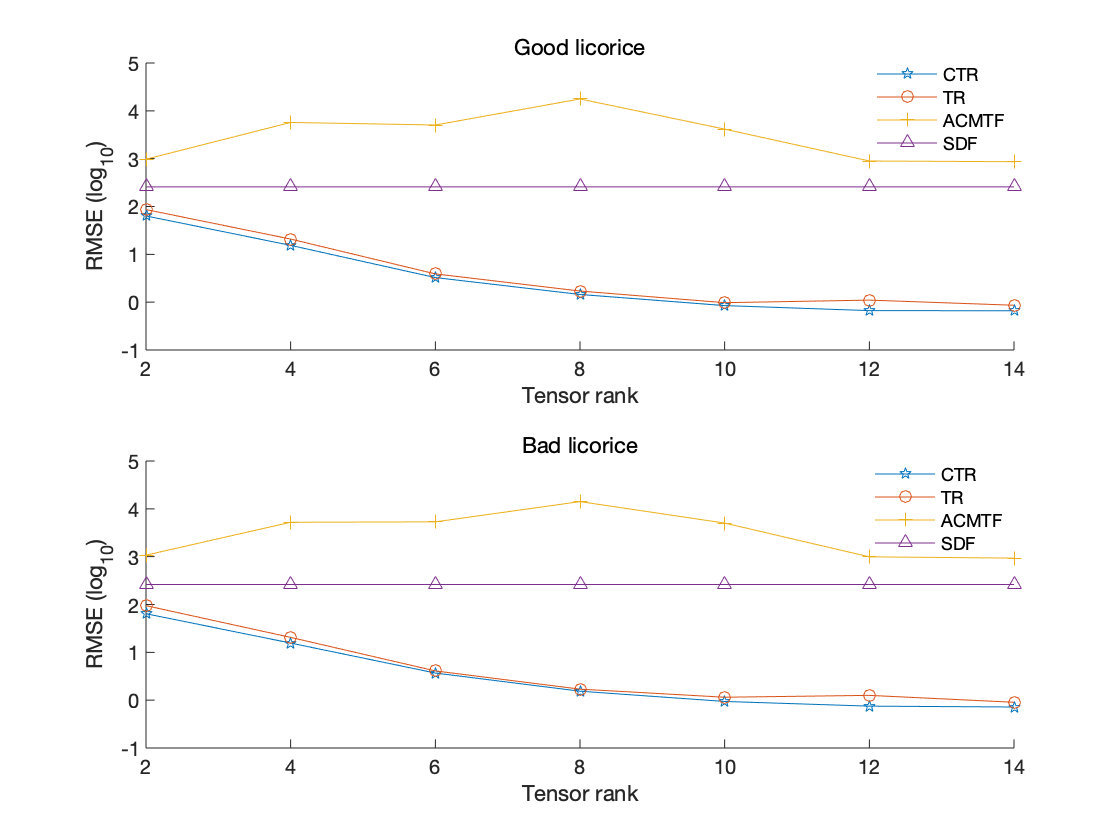}
\subcaption*{RMSE ($\log_{10}$)}
\end{subfigure}
\qquad\qquad
\begin{subfigure}[t]{0.45\textwidth}
\centering
\setlength{\abovecaptionskip}{0pt}
\setlength{\belowcaptionskip}{-2pt}
\includegraphics[scale=0.22]{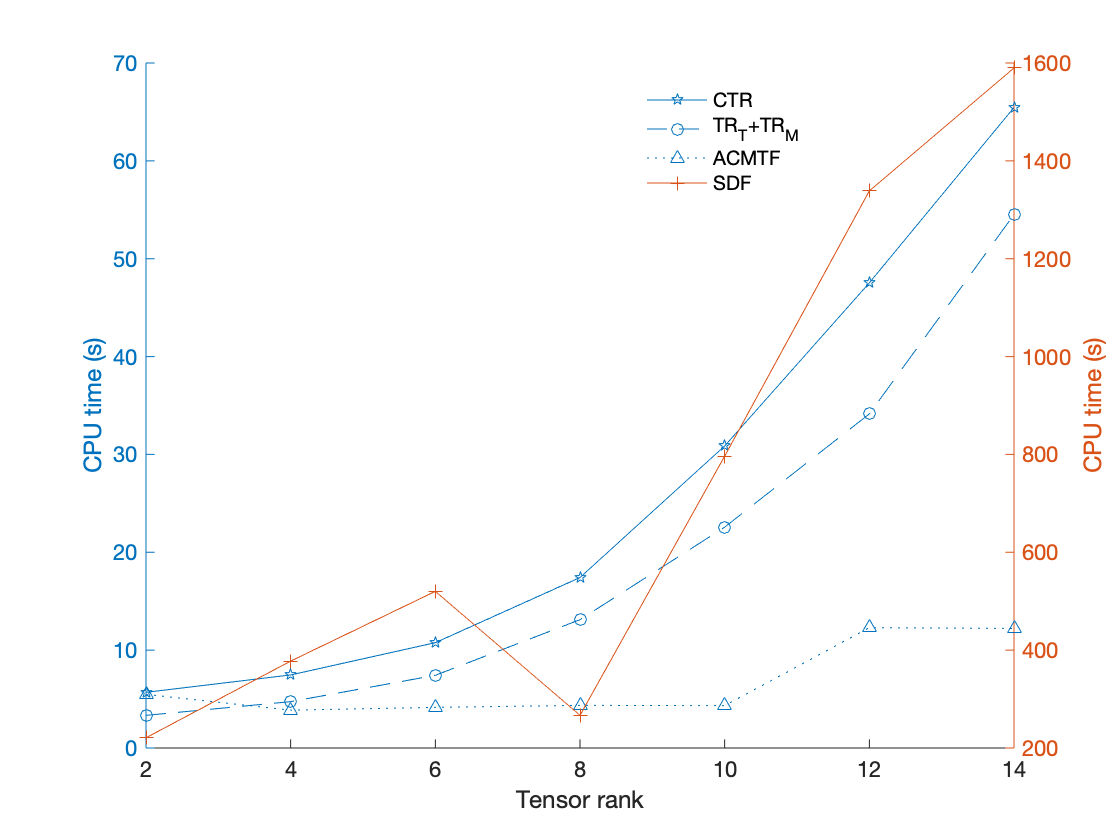}
\subcaption*{CPU time (s)}
\end{subfigure}
\caption{The completion result of the Licorice data derived by five algorithms. The panel (a) is the RMSE comparison and the panel (b) is the elapsed CPU time measured in seconds.}
\label{result-real-data3}
\end{figure*}

The completion result is shown in Fig. \ref{result-real-data3}. Note that the CNN method cannot process the scenario that both data dimensions are more than $3$, therefore we do not plot its result. From Fig. \ref{result-real-data3}(a), our CTRC method yields lower RMSE than its uncoupled version. Moreover, CTRC has lower RMSE than other coupled completion methods.

\section{Conclusion}
\label{section_conclusion}

This paper proposes to use tensor ring for coupled tensors completion, and a block coordinate descent algorithm is developed for its solution. We provide an excess risk bound for the proposed method, which implies the sampling complexity can be reduced to below the bound for individual tensor completion. The numerical results confirm the theoretical analysis and demonstrate the completion performance improvement in comparison with the state-of-the-art methods. Considering the high computational complexity of CTRC, the future work may focus on accelerating the algorithm for fast coupled completion.


%

\appendices
\section{Optimization on Coupled TR factors of $\mathfrak{R}_1$ and $\mathfrak{R}_2$}
\label{appendix_A}

To solve problem (\ref{model-tensor ring3_i_d}), we calculate the second-order partial derivatives of the objective function with respect to $\boldsymbol{\alpha}^{\left(d\right)}_{i_d}$, $\boldsymbol{\beta}^{\left(d\right)}_{i_d}$, $\boldsymbol{\gamma}^{\left(d\right)}_{i_d}$, respectively. We write the objective function as
\begin{align*}
f^{\left(d\right)}_{i_d}=& \frac{1}{2}\left[\boldsymbol{\alpha}^{\left(d\right)}_{i_d},\boldsymbol{\beta}^{\left(d\right)}_{i_d}\right]\widetilde{\mathbf{H}}^{\left(d\right)}_{i_d}\left[\boldsymbol{\alpha}^{\left(d\right)}_{i_d},\boldsymbol{\beta}^{\left(d\right)}_{i_d}\right]^{\mathrm{T}}- \\
& \left[\boldsymbol{\alpha}^{\left(d\right)}_{i_d},\boldsymbol{\beta}^{\left(d\right)}_{i_d}\right]\mathbf{P}^{\mathrm{T}}_d\overline{\mathbf{B}}_d\overline{\mathbf{c}}_{i_d}^{\left(d\right)^{\mathrm{T}}}+\frac{1}{2}\overline{\mathbf{c}}_{i_d}^{\left(d\right)}\overline{\mathbf{c}}_{i_d}^{\left(d\right)^{\mathrm{T}}}+ \\
& \frac{1}{2}\left[\boldsymbol{\alpha}^{\left(d\right)}_{i_d},\boldsymbol{\gamma}^{\left(d\right)}_{i_d}\right]\widetilde{\mathbf{H'}}^{\left(d\right)}_{i_d}\left[\boldsymbol{\alpha}^{\left(d\right)}_{i_d},\boldsymbol{\gamma}^{\left(d\right)}_{i_d}\right]^{\mathrm{T}}- \\
& \left[\boldsymbol{\alpha}^{\left(d\right)}_{i_d},\boldsymbol{\gamma}^{\left(d\right)}_{i_d}\right]\mathbf{P'}^{\mathrm{T}}_d\overline{\mathbf{B'}}_d\overline{\mathbf{c'}}_{i_d}^{\left(d\right)^{\mathrm{T}}}+\frac{1}{2}\overline{\mathbf{c'}}_{i_d}^{\left(d\right)}\overline{\mathbf{c'}}_{i_d}^{\left(d\right)^{\mathrm{T}}},
\end{align*}
then there is
\begin{align*}
& \frac{\partial{f_{i_d}}}{\partial{\boldsymbol{\alpha}^{\left(d\right)}_{i_d}}} \\
=& \frac{\partial}{\partial{\boldsymbol{\alpha}^{\left(d\right)}_{i_d}}}\left\{ \frac{1}{2}\boldsymbol{\alpha}^{\left(d\right)}_{i_d} \left(\widetilde{\mathbf{H}}_{i_d}^{\left(d\right)11}+\widetilde{\mathbf{H'}}_{i_d}^{\left(d\right)11}\right) \boldsymbol{\alpha}^{\left(d\right)^{\mathrm{T}}}_{i_d}+ \right. \\
& \boldsymbol{\alpha}^{\left(d\right)}_{i_d}\left[ \boldsymbol{\beta}^{\left(d\right)}_{i_d}\left( \widetilde{\mathbf{H}}_{i_d}^{\left(d\right)21}+\widetilde{\mathbf{H}}_{i_d}^{\left(d\right)12^{\mathrm{T}}} \right)+\boldsymbol{\gamma}^{\left(d\right)}_{i_d}\left( \widetilde{\mathbf{H'}}_{i_d}^{\left(d\right)21}+ \right. \right. \\
& \left. \left. \widetilde{\mathbf{H'}}_{i_d}^{\left(d\right)12^{\mathrm{T}}} \right)+\boldsymbol{\xi}^{\left(d\right)}_{i_d}+\boldsymbol{\xi'}^{\left(d\right)}_{i_d} \right]^{\mathrm{T}}+\frac{1}{2}\boldsymbol{\beta}^{\left(d\right)}_{i_d}\widetilde{\mathbf{H}}_{i_d}^{\left(d\right)22} \boldsymbol{\beta}^{\left(d\right)^{\mathrm{T}}}_{i_d}+ \\
& \frac{1}{2}\boldsymbol{\gamma}^{\left(d\right)}_{i_d}\widetilde{\mathbf{H'}}_{i_d}^{\left(d\right)22} \boldsymbol{\gamma}^{\left(d\right)^{\mathrm{T}}}_{i_d}+\boldsymbol{\beta}^{\left(d\right)}_{i_d}\boldsymbol{\eta}^{\left(d\right)^{\mathrm{T}}}_{i_d}+\boldsymbol{\gamma}^{\left(d\right)}_{i_d}\boldsymbol{\eta'}^{\left(d\right)^{\mathrm{T}}}_{i_d}+ \\
& \left. \frac{1}{2}z^{\left(d\right)}_{i_d}+\frac{1}{2}{z'}^{\left(d\right)}_{i_d} \right\} \\
=& \boldsymbol{\alpha}^{\left(d\right)}_{i_d} \left(\widetilde{\mathbf{H}}_{i_d}^{\left(d\right)11}+\widetilde{\mathbf{H'}}_{i_d}^{\left(d\right)11}\right)+\boldsymbol{\varrho}^{\left(d\right)}_{i_d},
\end{align*}
where
\begin{align*}
& \boldsymbol{\varrho}^{\left(d\right)}_{i_d} \\
=& \boldsymbol{\beta}^{\left(d\right)}_{i_d}\left( \widetilde{\mathbf{H}}_{i_d}^{\left(d\right)21}+\widetilde{\mathbf{H}}_{i_d}^{\left(d\right)12^{\mathrm{T}}} \right)+\boldsymbol{\gamma}^{\left(d\right)}_{i_d}\left( \widetilde{\mathbf{H'}}_{i_d}^{\left(d\right)21}+\widetilde{\mathbf{H'}}_{i_d}^{\left(d\right)12^{\mathrm{T}}} \right) \\
& +\boldsymbol{\xi}^{\left(d\right)}_{i_d}+\boldsymbol{\xi'}^{\left(d\right)}_{i_d}.
\end{align*}
Thus we have
\begin{equation}
\left\{
\begin{aligned}
& \frac{\partial^2{f_{i_d}}}{\partial{\boldsymbol{\alpha}^{\left(d\right)^2}_{i_d}}}=\widetilde{\mathbf{H}}_{i_d}^{\left(d\right)11}+\widetilde{\mathbf{H'}}_{i_d}^{\left(d\right)11} \\
& \frac{\partial^2{f_{i_d}}}{\partial{\boldsymbol{\alpha}^{\left(d\right)}_{i_d}}\partial{\boldsymbol{\beta}^{\left(d\right)}_{i_d}}}=\widetilde{\mathbf{H}}_{i_d}^{\left(d\right)12}+\widetilde{\mathbf{H}}_{i_d}^{\left(d\right)21^{\mathrm{T}}} \\
& \frac{\partial^2{f_{i_d}}}{\partial{\boldsymbol{\alpha}^{\left(d\right)}_{i_d}}\partial{\boldsymbol{\gamma}^{\left(d\right)}_{i_d}}}=\widetilde{\mathbf{H'}}_{i_d}^{\left(d\right)12}+\widetilde{\mathbf{H'}}_{i_d}^{\left(d\right)21^{\mathrm{T}}}
\end{aligned}
\right..
\label{Hessian1}
\end{equation}

Then we deduce
\begin{align*}
\frac{\partial{f_{i_d}}}{\partial{\boldsymbol{\beta}^{\left(d\right)}_{i_d}}}=& \frac{\partial}{\partial{\boldsymbol{\beta}^{\left(d\right)}_{i_d}}}\left\{ \frac{1}{2}\boldsymbol{\beta}^{\left(d\right)}_{i_d}\widetilde{\mathbf{H}}_{i_d}^{\left(d\right)22}\boldsymbol{\beta}^{\left(d\right)^{\mathrm{T}}}_{i_d}+\boldsymbol{\beta}^{\left(d\right)}_{i_d}\left[ \boldsymbol{\alpha}^{\left(d\right)}_{i_d}\left( \widetilde{\mathbf{H}}_{i_d}^{\left(d\right)12} \right. \right. \right. \\
& \left. \left. +\widetilde{\mathbf{H}}_{i_d}^{\left(d\right)21^{\mathrm{T}}} \right)+\boldsymbol{\eta}^{\left(d\right)}_{i_d} \right]^{\mathrm{T}}+\frac{1}{2}\boldsymbol{\gamma}^{\left(d\right)}_{i_d}\widetilde{\mathbf{H'}}_{i_d}^{\left(d\right)22} \boldsymbol{\gamma}^{\left(d\right)^{\mathrm{T}}}_{i_d}+ \\
& \boldsymbol{\gamma}^{\left(d\right)}_{i_d}\left[ \boldsymbol{\alpha}^{\left(d\right)}_{i_d} \left( \widetilde{\mathbf{H'}}_{i_d}^{\left(d\right)12}+\widetilde{\mathbf{H'}}_{i_d}^{\left(d\right)21^{\mathrm{T}}} \right)+\boldsymbol{\eta'}^{\left(d\right)}_{i_d} \right]^{\mathrm{T}}+ \\
& \frac{1}{2}\boldsymbol{\alpha}^{\left(d\right)}_{i_d} \left(\widetilde{\mathbf{H}}_{i_d}^{\left(d\right)11}+\widetilde{\mathbf{H'}}_{i_d}^{\left(d\right)11}\right) \boldsymbol{\alpha}^{\left(d\right)^{\mathrm{T}}}_{i_d}+ \\
& \left. \boldsymbol{\alpha}^{\left(d\right)}_{i_d}\left(\boldsymbol{\xi}^{\left(d\right)}_{i_d}+\boldsymbol{\xi'}^{\left(d\right)}_{i_d}\right)^{\mathrm{T}}+\frac{1}{2}z^{\left(d\right)}_{i_d}+\frac{1}{2}{z'}^{\left(d\right)}_{i_d} \right\} \\
=& \boldsymbol{\beta}^{\left(d\right)}_{i_d}\widetilde{\mathbf{H}}_{i_d}^{\left(d\right)22}+\boldsymbol{\vartheta}^{\left(d\right)}_{i_d},
\end{align*}
where $\boldsymbol{\vartheta}^{\left(d\right)}_{i_d}=\boldsymbol{\alpha}^{\left(d\right)}_{i_d}\left( \widetilde{\mathbf{H}}_{i_d}^{\left(d\right)12}+\widetilde{\mathbf{H}}_{i_d}^{\left(d\right)21^{\mathrm{T}}} \right)+\boldsymbol{\eta}^{\left(d\right)}_{i_d}$. Hence there is
\begin{equation}
\left\{
\begin{aligned}
& \frac{\partial^2{f_{i_d}}}{\partial{\boldsymbol{\beta}^{\left(d\right)^2}_{i_d}}}=\widetilde{\mathbf{H}}_{i_d}^{\left(d\right)22},\; \frac{\partial^2{f_{i_d}}}{\partial{\boldsymbol{\beta}^{\left(d\right)}_{i_d}}\partial{\boldsymbol{\alpha}^{\left(d\right)}_{i_d}}}=\widetilde{\mathbf{H}}_{i_d}^{\left(d\right)21}+\widetilde{\mathbf{H}}_{i_d}^{12^{\mathrm{T}}} \\
& \frac{\partial^2{f_{i_d}}}{\partial{\boldsymbol{\beta}^{\left(d\right)}_{i_d}}\partial{\boldsymbol{\gamma}^{\left(d\right)}_{i_d}}}=\mathbf{0}
\end{aligned}
\right..
\label{Hessian2}
\end{equation}
Similarly, we derive
\begin{equation}
\left\{
\begin{aligned}
& \frac{\partial^2{f_{i_d}}}{\partial{\boldsymbol{\gamma'}^{\left(d\right)^2}_{i_d}}}=\widetilde{\mathbf{H'}}_{i_d}^{\left(d\right)22},\; \frac{\partial^2{f_{i_d}}}{\partial{\boldsymbol{\gamma}^{\left(d\right)}_{i_d}}\partial{\boldsymbol{\alpha}^{\left(d\right)}_{i_d}}}=\widetilde{\mathbf{H'}}_{i_d}^{\left(d\right)21}+\widetilde{\mathbf{H'}}_{i_d}^{\left(d\right)12^{\mathrm{T}}} \\
& \frac{\partial^2{f_{i_d}}}{\partial{\boldsymbol{\gamma}^{\left(d\right)}_{i_d}}\partial{\boldsymbol{\beta}^{\left(d\right)}_{i_d}}}=\mathbf{0}
\end{aligned}
\right..
\label{Hessian3}
\end{equation}

Incorporating (\ref{Hessian1}) -- (\ref{Hessian3}), we derive the Hessian matrix
\begin{align*}
& \widehat{\mathbf{H}}_{i_d}= \\
& \begin{bmatrix}
\widetilde{\mathbf{H}}_{i_d}^{\left(d\right)11}+\widetilde{\mathbf{H'}}_{i_d}^{\left(d\right)11} & \widetilde{\mathbf{H}}_{i_d}^{\left(d\right)12}+\widetilde{\mathbf{H}}_{i_d}^{\left(d\right)21^{\mathrm{T}}} & \widetilde{\mathbf{H'}}_{i_d}^{\left(d\right)12}+\widetilde{\mathbf{H'}}_{i_d}^{\left(d\right)21^{\mathrm{T}}} \\
\widetilde{\mathbf{H}}_{i_d}^{\left(d\right)21}+\widetilde{\mathbf{H}}_{i_d}^{\left(d\right)12^{\mathrm{T}}} & \widetilde{\mathbf{H}}_{i_d}^{\left(d\right)22} & \mathbf{0} \\
\widetilde{\mathbf{H'}}_{i_d}^{\left(d\right)21}+\widetilde{\mathbf{H'}}_{i_d}^{\left(d\right)12^{\mathrm{T}}} & \mathbf{0} & \widetilde{\mathbf{H'}}_{i_d}^{\left(d\right)22}
\end{bmatrix}
.
\end{align*}

\section{Proof of Theorem \ref{theorem_risk bound}}
\label{appendix_B}

\subsection{The expectation of a linear combination of products of independent variables}

Given a non-standardized Student's t variable $X$ with parameters $a$ and $b$, then the density function of $Y=X^2$ is 
\begin{align*}
p\left(y;a,b\right)=\frac{\Gamma\left(a+\frac{1}{2}\right)}{\sqrt{2\pi b}\Gamma\left(a\right)}y^{-\frac{1}{2}}\left(1+\frac{y}{2b}\right)^{-\left(a+\frac{1}{2}\right)}.
\end{align*}
 We recognize it as a non-standardized Fisher–Snedecor (we use F below for conciseness) distribution with degrees of freedom $1$, $2a$ and scale $b/a$.

The PDF of the sum of F random variables is given by incorporating the parameters $a$ and $b$ with \cite{du2020sum}:
\begin{align*}
p\left(z\right)=\frac{z^{-1}}{\pi^{\frac{k}{2}}\Gamma^k\left(a\right)}H_{\mathrm{PDF}},
\end{align*}
where the multivariable Fox’s H-function is defined as
\begin{align*}
H_{\mathrm{PDF}}\triangleq H^{0,0:1,2;1,2;\dotsc;1,2}_{0,1:2,1;2,1;\dotsc;2,1}\left(\substack{ -: \left(1,1\right),\left(1-a\right);\dotsc,;\left(1,1\right),\left(1-a,1\right) \\ \left(1;1,\dotsc,1\right):\left(\frac{1}{2},1\right);\dotsc;\left(\frac{1}{2},1\right) } \Bigg| \substack{ \frac{z}{2b} \\ \vdots \\ \frac{z}{2b} } \right).
\end{align*}
It has an alternative expression
\begin{align*}
p\left(z\right)=& \frac{z^{-1}}{\pi^{\frac{k}{2}}\Gamma^k\left(a\right)}\frac{1}{\left(2\pi \mathfrak{i}\right)^k} \\
& \int_{\mathcal{L}_1}\dotsm \int_{\mathcal{L}_k} \frac{1}{\Gamma\left(\zeta\right)} \prod^{k}_{t=1}\Upsilon\left(\zeta_t\right) \left(\frac{z}{2b}\right)^{\zeta_t} d\zeta_1 \dotsm \zeta_k,
\end{align*}
where $\zeta=\sum^{k}_{t=1}\zeta_t$ and $\Upsilon\left(\zeta_t\right)=\Gamma\left(\frac{1}{2}-\zeta_t\right)\Gamma\left(\zeta_t\right)\Gamma\left(a+\zeta_t\right)$.

Supposing $X_i$, $i=1, \dotsc, m$ and $Y_j$, $j=1, \dotsc, n$ are the independent non-standardized Fisher–Snedecor random variables with the same parameters $a$ and $b$. We assume $0<\alpha<1$ and $0<\beta<1$, then the expectation of $\sqrt{\alpha \prod^m_{i=1}X_i+\beta \prod^n_{j=1}Y_j}$ is given by the multiple integral
\begin{align}
& \int^{+\infty}_0\dotsm \int^{+\infty}_0 \sqrt{\alpha \prod^m_{i=1}x_i+\beta \prod^n_{j=1}y_j}\prod^m_{i=1} p\left(x_i\right)\prod^n_{j=1}p\left(y_j\right) \notag \\
& \mathrm{d}x_1\dotsm \mathrm{d}x_m \mathrm{d}y_1\dotsm \mathrm{d}y_n.
\label{integral-1}
\end{align}

The calculation of this integral is done with the help of the method of brackets, which expands a definite integral evaluating over the half line $\left[0,+\infty\right)$ as a series consisting of the brackets. For example, the notation $\langle a \rangle$ stands for the divergent integral $\int^{+\infty}_0 x^{a-1}\mathrm{d}x$. If function $f\left(x\right)$ admits the formal power series $\sum^{+\infty}_{n=0}a_nx^{c n+d-1}$, then the improper integral of $f$ is formalized by $\int^{+\infty}_0f\left(x\right)\mathrm{d}x=\sum^{+\infty}_{n=0}a_n\langle c n+d \rangle$. The indicator $\phi_n\triangleq \left(-1\right)^n/\Gamma\left(n+1\right)$ will be used in the series expressions when applying the method of brackets. The Pochhammer symbols defined as $\left(b\right)_n\triangleq \Gamma\left(n+b\right)/\Gamma\left(b\right)$ is a systematic procedure in the simplification of the series. An exponential function $\exp\left(-x\right)$ can be represented as $\sum_n \phi_n x^n$ in the framework of the method of brackets. Another useful rule is that a multinomial $\left(x_1+\dotsm+x_m\right)^a$ is expanded as $\sum_{\left\{n\right\}}\phi_{\left\{n\right\}}x_1^{n_1}\dotsm x_m^{n_m}\langle n_1+\dotsm+n_m-a \rangle/\Gamma\left(-a\right)$.

The following equations follow from the brackets method and the final value theorem in Laplace transform will be useful:
\begin{align}
\langle o \rangle=\lim\limits_{s\to 0} \frac{\Gamma\left(o+1\right)}{s^{o+1}}.
\label{equation-1}
\end{align}
Incorporating (\ref{equation-1}) along with the Fubini’s theorem we have
\begin{align}
& \mathbb{E}\left[X_i^p\right]=\lim\limits_{s\to 0} \frac{1}{s^p} \frac{1}{\pi^{\frac{k}{2}}\Gamma^k\left(a\right)} \frac{1}{\left(2\pi \mathfrak{i}\right)^k} \notag \\
& \int_{\mathcal{L}^{\left(i\right)}_1}\dotsm\int_{\mathcal{L}^{\left(i\right)}_k} \left(\zeta^{\left(i\right)}\right)_p \prod^{k}_{t=1}\Upsilon\left(\zeta^{\left(i\right)}_t\right) \left(\frac{1}{2bs}\right)^{\zeta^{\left(i\right)}_t} \mathrm{d}\zeta^{\left(i\right)}_1 \dotsm \mathrm{d}\zeta^{\left(i\right)}_k \notag \\
& =\lim\limits_{s\to 0} \frac{1}{s^p} \frac{1}{\pi^{\frac{k}{2}}\Gamma^k\left(a\right)} \notag \\
& H^{0,1:1,2;1,2;\dotsc;1,2}_{1,1:2,1;2,1;\dotsc;2,1}\left(\substack{ \left(1-p;1,\dotsc,1\right): \left(1,1\right),\left(1-a\right);\dotsc,;\left(1,1\right),\left(1-a,1\right) \\ \left(1;1,\dotsc,1\right):\left(\frac{1}{2},1\right);\dotsc;\left(\frac{1}{2},1\right) } \Bigg| \substack{ \frac{1}{2bs} \\ \vdots \\ \frac{1}{2bs} } \right)
\label{equation-2_1}
\end{align}
and
\begin{align}
& \mathbb{E}\left[Y_j^q\right]=\lim\limits_{s\to 0} \frac{1}{s^q} \frac{1}{\pi^{\frac{k}{2}}\Gamma^k\left(a\right)} \frac{1}{\left(2\pi \mathfrak{i}\right)^k} \notag \\
& \int_{\mathcal{L}^{\left(j\right)}_1}\dotsm\int_{\mathcal{L}^{\left(j\right)}_k} \left(\eta^{\left(j\right)}\right)_q \prod^{k}_{t=1}\Upsilon\left(\eta^{\left(j\right)}_t\right) \left(\frac{1}{2bs}\right)^{\eta^{\left(j\right)}_t} \mathrm{d}\eta^{\left(j\right)}_1 \dotsm \mathrm{d}\eta^{\left(j\right)}_k \notag \\
& =\lim\limits_{s\to 0} \frac{1}{s^q} \frac{1}{\pi^{\frac{k}{2}}\Gamma^k\left(a\right)} \notag \\
& H^{0,1:1,2;1,2;\dotsc;1,2}_{1,1:2,1;2,1;\dotsc;2,1}\left(\substack{ \left(1-q;1,\dotsc,1\right): \left(1,1\right),\left(1-a\right);\dotsc,;\left(1,1\right),\left(1-a,1\right) \\ \left(1;1,\dotsc,1\right):\left(\frac{1}{2},1\right);\dotsc;\left(\frac{1}{2},1\right) } \Bigg| \substack{ \frac{1}{2bs} \\ \vdots \\ \frac{1}{2bs} } \right),
\label{equation-2_2}
\end{align}
where $\zeta^{\left(i\right)}=\sum^{k}_{t=1} \zeta^{\left(i\right)}_t$, $\eta^{\left(j\right)}=\sum^{k}_{t=1} \eta^{\left(j\right)}_t$. We use $H_{X_i}\left(1-p,\sim \big| \mathbf{\frac{1}{2bs}}\right)$ and $H_{Y_j}\left(1-q,\sim \big| \mathbf{\frac{1}{2bs}}\right)$ as the brief notes of two Fox's H-functions. The boldface $\mathbf{1/2bs}$ is used to represent the vector input of the multivariate Fox's H-function for simplicity.

In order to solve (\ref{integral-1}), we start with (\ref{equation-2_1}), (\ref{equation-2_2}) and the brackets method. Slinging out the terms that do not contain the integral variables, merging the remained terms and substituting the integral with brackets, and the integral (\ref{integral-1}) is transformed into
\begin{align}
& \frac{\lim\limits_{s\to 0}}{\Gamma\left(-\frac{1}{2}\right)\pi^{\frac{k\left(m+n\right)}{2}}\Gamma^{k\left(m+n\right)}\left(a\right)} \sum^{+\infty}_{l_1,l_2=0}\phi_{l_1l_2}\alpha^{l_1}\beta^{l_2} \langle l_1+l_2-\frac{1}{2} \rangle \notag \\
& \frac{1}{s^{ml_1}} H^{m}_{X}\left(1-l_1,\sim \big| \mathbf{\frac{1}{2bs}}\right) \frac{1}{s^{nl_2}} H^{n}_{Y}\left(1-l_2,\sim \big| \mathbf{\frac{1}{2bs}}\right)
\label{integral-2}
\end{align}
due to the independency of $X_i$ and $Y_j$.

Afterwards, the matrix of coefficients left has rank $1$, thus it produces two series as candidates for the values of the integral, one per free variable. We choose $l_1$ or $l_2$ as a free variable and eliminate the bracket. The result shown below follows from the rule that the value assigned to $\sum_n \phi_n f\left(n\right)\langle cn+d \rangle$ is $f\left(n^*\right)\Gamma\left(-n^*\right)/\left| c \right|$, where $n^*$ is obtained from the vanishing of the brackets. The simplification of (\ref{integral-2}) is given as
\begin{align}
& \frac{\sqrt{\beta} \lim\limits_{s\to 0}}{\Gamma\left(-\frac{1}{2}\right)\pi^{\frac{k\left(m+n\right)}{2}}\Gamma^{k\left(m+n\right)}\left(a\right)} \sum^{+\infty}_{l=0} \frac{\left(-\frac{\alpha}{\beta}\right)^{l}}{l!} \notag \\
& \frac{1}{s^{ml}} H^{m}_{X}\left(1-l,\sim \big| \mathbf{\frac{1}{2bs}}\right) \frac{1}{s^{n \left(\frac{1}{2}-l\right)}} H^{n}_{Y}\left(\frac{1}{2}+l,\sim \big| \mathbf{\frac{1}{2bs}}\right)
\label{integral-3_1}
\end{align}
or
\begin{align}
& \frac{\sqrt{\alpha} \lim\limits_{s\to 0}}{\Gamma\left(-\frac{1}{2}\right)\pi^{\frac{k\left(m+n\right)}{2}}\Gamma^{k\left(m+n\right)}\left(a\right)} \sum^{+\infty}_{l=0} \frac{\left(-\frac{\beta}{\alpha}\right)^{l}}{l!} \notag \\
& \frac{1}{s^{m \left(\frac{1}{2}-l\right)}} H^{m}_{X}\left(\frac{1}{2}+l,\sim \big| \mathbf{\frac{1}{2bs}}\right) \frac{1}{s^{nl}} H^{n}_{Y}\left(1-l,\sim \big| \mathbf{\frac{1}{2bs}}\right).
\label{integral-3_2}
\end{align}

However, this result is insufficient for the analysis. we utilize an accurate closed-form approximation introduced in \cite{du2020sum}, which approximates a sum of F random variables using a single F random variable. The key idea is to introduce a variable to parameterize the parameters of a F distribution, say $\varepsilon$, which can be tuned to minimize the Kolmogorov distance using the moment matching method.

We designate $\mathcal{F}\left(df_1\left(\varepsilon\right),df_2\left(\varepsilon\right)\right)$ as the shorthand for a standard F distribution. Its PDF is
\begin{align*}
p\left(z\right)=\frac{{df_1\left(\varepsilon\right)}^{df_1\left(\varepsilon\right)}{df_2\left(\varepsilon\right)}^{df_2\left(\varepsilon\right)}x^{df_1\left(\varepsilon\right)-1}}{\operatorname{B}\left(df_1\left(\varepsilon\right),df_2\left(\varepsilon\right)\right)\left(df_1\left(\varepsilon\right)x+df_2\left(\varepsilon\right)\right)^{df_1\left(\varepsilon\right)+df_2\left(\varepsilon\right)}},
\end{align*}
where the functions $df_1\left(\varepsilon\right)$ and $df_2\left(\varepsilon\right)$ are defined by (\ref{equation-3_1}) and (\ref{equation-3_2}) respectively. The optimal values of $df_1\left(\varepsilon^*\right)$ and $df_2\left(\varepsilon^*\right)$ are noted as $df_1^*$ and $df_2^*$. The PDF of the non-standardized one obeys $(a/b)p\left(az/b\right)$.
\begin{figure*}[!t]
\setcounter{equation}{30}
\begin{align}
& df_1\left(\varepsilon\right)= \notag \\
& k\frac{-\left(a-2\right)^2\left(a-3\right)\varepsilon^2 -\left(a-2\right)\left[ a^2\left(k-2\right)-a\left(5k-14\right)+6k-14 \right]\varepsilon + \left(2a-1\right)\left[ a^2\left(k+2\right)-a\left(5k+1\right)+6k-3 \right]}{\left(a-2\right)^2\left(2ak+a-6k+2\right)\varepsilon^2 + \left(a-2\right)\left[ a^2\left(7k+6\right)-28ak+16k+a-2 \right]\varepsilon +2\left(2a-1\right)\left[ a^2\left(k+2\right)-a\left(5k+1\right)+3k \right]}
\label{equation-3_1}
\end{align}
\begin{align}
& df_2\left(\varepsilon\right)= \notag \\
& \frac{\left(a-2\right)^2\left(a-3\right)\varepsilon^2 + \left(a-2\right)\left[ a^2\left(k-5\right) - a\left(5k-28\right) + 6\left(k-4\right) \right]\varepsilon - \left(2a-1\right)\left[ a^2\left(k+2\right) - a\left(5k-2\right) + 6\left(k-1\right) \right] }{\left(a-2\right)^2\left(a-3\right)\varepsilon^2 - \left(3a^3-20a^2+38a-20\right)\varepsilon - 6\left(a-\frac{1}{2}\right)\left(a-1\right)}
\label{equation-3_2}
\end{align}
\hrulefill
\end{figure*}

Following the similar mathematical manipulation above, (\ref{integral-1}) is rewritten as
\begin{align}
& \int^{+\infty}_0\dotsm \int^{+\infty}_0 \frac{1}{\Gamma\left(-\frac{1}{2}\right)} \sum^{+\infty}_{l_1,l_2=0}\phi_{l_1l_2}\alpha^{l_1}\beta^{l_2} \langle l_1+l_2-\frac{1}{2} \rangle \notag \\
& \prod^{m}_{i=1}\frac{a{df^*_1}^{df^*_1} {df^*_2}^{df^*_2} \left(\frac{a}{b}x_i\right)^{df^*_1-1}}{b\operatorname{B}\left(df^*_1,df^*_2\right) \left(df^*_1\frac{a}{b}x_i+df^*_2\right)^{df^*_1+df^*_2}} \notag \\
& \prod^{n}_{j=1}\frac{a{df^*_1}^{df^*_1} {df^*_2}^{df^*_2} \left(\frac{a}{b}y_j\right)^{df^*_1-1}}{b\operatorname{B}\left(df^*_1,df^*_2\right) \left(df^*_1\frac{a}{b}y_j+df^*_2\right)^{df^*_1+df^*_2}} \notag \\
& \mathrm{d}x_1\dotsm \mathrm{d}x_m \mathrm{d}y_1\dotsm \mathrm{d}y_n.
\label{integral-4}
\end{align}
Expanding $\left(df^*_1\frac{a}{b}x_i+df^*_2\right)^{df^*_1+df^*_2}$ as
\begin{align*}
\sum^{+\infty}_{p_{i1},p_{i2}=0} \phi_{p_{i1}p_{i2}} \left(df^*_1\frac{a}{b}x_i\right)^{p_{i1}} \left(df^*_2\right)^{p_{i2}} \frac{\langle p_{i1}+p_{i2}+df^*_1+df^*_2 \rangle}{\Gamma\left(df^*_1+df^*_2\right)},
\end{align*}
and $\left(df^*_1\frac{a}{b}y_j+df^*_2\right)^{df^*_1+df^*_2}$ as
\begin{align*}
\sum^{+\infty}_{q_{i1},q_{i2}=0} \phi_{q_{i1}q_{i2}} \left(df^*_1\frac{a}{b}y_j\right)^{q_{i1}} \left(df^*_2\right)^{q_{i2}} \frac{\langle q_{i1}+q_{i2}+df^*_1+df^*_2 \rangle}{\Gamma\left(df^*_1+df^*_2\right)},
\end{align*}
then collecting the terms that contain $x_i$ and $y_j$ respectively, substituting the integrals with the brackets and slinging out the constant terms there is
\begin{align}
& \frac{1}{\Gamma\left(-\frac{1}{2}\right)} \sum^{+\infty}_{l_1,l_2=0}\phi_{l_1l_2} \langle l_1+l_2-\frac{1}{2} \rangle \notag \\
& \frac{\Gamma^{m}\left(df^*_1+l_1\right)\Gamma^{m}\left(df^*_2-l_1\right)}{\Gamma^{m}\left(df^*_1\right)\Gamma^{m}\left(df^*_2\right)} \left[\alpha\left(\frac{df^*_2b}{df^*_1a}\right)^{m}\right]^{l_1} \notag \\
& \frac{\Gamma^{n}\left(df^*_1+l_2\right)\Gamma^{n}\left(df^*_2-l_2\right)}{\Gamma^{n}\left(df^*_1\right)\Gamma^{n}\left(df^*_2\right)} \left[\beta\left(\frac{df^*_2b}{df^*_1a}\right)^{n}\right]^{l_2}.
\label{integral-5}
\end{align}
Next we use a transformed Pochhammer symbol $\left(b\right)_{-n}=\left(-1\right)^n/\left(1-b\right)_n$ which can be proved by repeating the recurrence relation $\Gamma\left(x\right)=\Gamma\left(x+1\right)/x=\left(x-1\right)\Gamma\left(x-1\right)$. By introducing the hypergeometric function $_PF_Q\left(\substack{\dotsm \\ \dotsm} | \cdot\right)$, the final simplifications of (\ref{integral-5}) hence are derived.
\subsubsection{Case $1$}
The variable $l_1$ is free. Thus plugging $l_2^*=1/2-l_1=l$ into the rule gives
\begin{align}
& \sqrt{\beta}\left(\frac{df^*_2b}{df^*_1a}\right)^{\frac{n}{2}} \sum^{+\infty}_{l=0} \frac{\left(-\frac{1}{2}\right)_{l} \left(df^*_1\right)^{m}_{l} \left(df^*_2-\frac{1}{2}\right)^{n}_{l}}{\left(1-df^*_2\right)^{m}_{l} \left(\frac{1}{2}-df^*_1\right)^{n}_{l}} \notag \\
& \frac{1}{l!} \left[ \left(-1\right)^{D_1+D_2+1}\frac{\alpha}{\beta}\left(\frac{df^*_2b}{df^*_1a}\right)^{D_1-D_2} \right]^{l} \notag \\
=& \sqrt{\beta}\left(\frac{df^*_2b}{df^*_1a}\right)^{\frac{n}{2}} {_{D_1+D_2-2L+1}}F_{D_1+D_2-2L} \notag \\
& \left( \substack{df^*_1,\dotsc,df^*_2-\frac{1}{2},\dotsc,-\frac{1}{2} \\ 1-df^*_2,\dotsc,\frac{1}{2}-df^*_1,\dotsc} \Bigg| \left(-1\right)^{D_1+D_2+1}\frac{\alpha}{\beta}\left(\frac{df^*_2b}{df^*_1a}\right)^{D_1-D_2} \right).
\label{integral-6_1}
\end{align}

\subsubsection{Case $2$}
The variable $l_2$ is free. Then plugging $l_1^*=1/2-l_2$ into the rule yields
\begin{align}
& \sqrt{\alpha}\left(\frac{df^*_2b}{df^*_1a}\right)^{\frac{m}{2}} \sum^{+\infty}_{l=0} \frac{\left(-\frac{1}{2}\right)_{l} \left(df^*_1\right)^{n}_{l} \left(df^*_2-\frac{1}{2}\right)^{m}_{l}}{\left(1-df^*_2\right)^{n}_{l} \left(\frac{1}{2}-df^*_1\right)^{m}_{l}} \notag \\
& \frac{1}{l!} \left[ \left(-1\right)^{D_1+D_2+1}\frac{\beta}{\alpha}\left(\frac{df^*_2b}{df^*_1a}\right)^{D_2-D_1} \right]^{l} \notag \\
=& \sqrt{\alpha}\left(\frac{df^*_2b}{df^*_1a}\right)^{\frac{m}{2}} {_{D_1+D_2-2L+1}}F_{D_1+D_2-2L} \notag \\
& \left( \substack{df^*_2-\frac{1}{2},\dotsc,df^*_1,\dotsc,-\frac{1}{2} \\ \frac{1}{2}-df^*_1,\dotsc,1-df^*_2,\dotsc} \Bigg| \left(-1\right)^{D_1+D_2+1}\frac{\beta}{\alpha}\left(\frac{df^*_2b}{df^*_1a}\right)^{D_2-D_1} \right).
\label{integral-6_2}
\end{align}
The selection of the two expressions depends on the convergence condition of the hypergeometric function. The first one is employed if the absolute value of the input variable of $F$ function is less than $1$ \cite{gonzalez2010method}, otherwise the second one is considered.

\subsection{Bounding the expectation of the $\mathrm{F}$-norm of coupled tensors}
Without loss of generality, supposing $\mathcal{X}$ and $\mathcal{Y}$ are $D_1$-order and $D_2$-order tensors coupled on their first $L$ modes, with the same TR rank and dimensional size. To calculate $\mathbb{E} \sqrt{\alpha \left\|\mathcal{X}\right\|^2_{\mathrm{F}}+\beta \left\|\mathcal{Y}\right\|^2_{\mathrm{F}}}$, we first note that $\left\|\cdot\right\|_{\mathrm{F}}$ is submultiplicative and the independency of the random tensors, thus using (\ref{integral-6_1}) we have
\begin{align}
& \mathbb{E} \sqrt{\alpha \left\|\mathcal{X}\right\|^2_{\mathrm{F}}+\beta \left\|\mathcal{Y}\right\|^2_{\mathrm{F}}} \notag \\
\leq & \mathbb{E} \prod^L_{l=1}\left\|\mathcal{U}^{\left(l\right)}\right\|_{\mathrm{F}} \sqrt{\alpha \prod^{D_1}_{d_1=L+1}\left\|\mathcal{U}^{\left(d_1\right)}\right\|^2_{\mathrm{F}}+\beta \prod^{D_2}_{d_2=L+1}\left\|\mathcal{V}^{\left(d_2\right)}\right\|^2_{\mathrm{F}}} \notag \\
=& \sqrt{\beta} \left(\frac{df^*_2b}{df^*_1a}\right)^{\frac{D_2}{2}} \frac{\operatorname{B}^{L}\left(df^*_1+\frac{1}{2},df^*_2-\frac{1}{2}\right)}{\operatorname{B}^{L}\left(df^*_1,df^*_2\right)} \notag \\
& {_{D_1+D_2-2L+1}}F_{D_1+D_2-2L} \notag \\
& \left( \substack{df^*_1,\dotsc,df^*_2-\frac{1}{2},\dotsc,-\frac{1}{2} \\ 1-df^*_2,\dotsc,\frac{1}{2}-df^*_1,\dotsc} \Bigg| \left(-1\right)^{D_1+D_2+1}\frac{\alpha}{\beta}\left(\frac{df^*_2b}{df^*_1a}\right)^{D_1-D_2} \right).
\label{bound1}
\end{align}
and using (\ref{integral-6_2}) we get
\begin{align}
& \mathbb{E} \sqrt{\alpha \left\|\mathcal{X}\right\|^2_{\mathrm{F}}+\beta \left\|\mathcal{Y}\right\|^2_{\mathrm{F}}} \notag \\
\leq& \sqrt{\alpha}\left(\frac{df^*_2b}{df^*_1a}\right)^{\frac{D_1}{2}} \frac{\operatorname{B}^{L}\left(df^*_1+\frac{1}{2},df^*_2-\frac{1}{2}\right)}{\operatorname{B}^{L}\left(df^*_1,df^*_2\right)} \notag \\
& {_{D_1+D_2-2L+1}}F_{D_1+D_2-2L} \notag \\
& \left( \substack{df^*_2-\frac{1}{2},\dotsc,df^*_1,\dotsc,-\frac{1}{2} \\ \frac{1}{2}-df^*_1,\dotsc,1-df^*_2,\dotsc} \Bigg| \left(-1\right)^{D_1+D_2+1}\frac{\beta}{\alpha}\left(\frac{df^*_2b}{df^*_1a}\right)^{D_2-D_1} \right).
\label{bound2}
\end{align}

\subsection{Bounding the excess risk}

A subset $\mathbf{x}_{m_1}$ containing $m_1=\left| \mathbb{S}_1 \cup \mathbb{T}_1 \right|$ elements is sampled uniformly without replacement from $\operatorname{vec}\left(\mathcal{X}\right)$. We concatenate $\mathbf{x}_{m_1}$ and $\mathbf{y}_{m_2}$ as a vector $\mathbf{z}_m\triangleq \left[\mathbf{x}_{m_1};\mathbf{y}_{m_2}\right]$ where $m=\left| \mathbb{S} \cup \mathbb{T} \right|$,
\begin{align*}
\hat{Q}_{m,n}\left(\bar{l}_{\mathbb{T}},\mathbf{z}_m\right)=\underset{\mathbf{z}_n}{\operatorname{E}}\left[ \sup_{\mathcal{X},\mathcal{Y}\in \mathcal{H}}\bar{l}_{\mathbb{T}}\left(\mathbf{z}_k,\mathbf{t}_k\right)-\bar{l}_{\mathbb{T}}\left(\mathbf{z}_n,\mathbf{t}_n\right) \right],
\end{align*}
where $\mathbf{z}_n$, $n\in \left\{1,\dotsc,m-1\right\}$ is a random subset of $\mathbf{z}_m$ containing $n$ elements sampled uniformly without replacement and $\mathbf{z}_k \triangleq \mathbf{z}_m\backslash \mathbf{z}_n$.

Under the hypothesis $\mathcal{H}$ mentioned before, letting $m=2n=\left| \mathbb{T}_1\cup \mathbb{T}_2 \right|$,  the expectation of the permutational Rademacher complexity is bounded as follows:
\begin{align*}
& \underset{\mathbf{z}_m}{\mathbb{E}}\left[ \hat{Q}_{m,m/2}\left(\bar{l}_{\mathbb{T}},\mathbf{z}_m\right) \right] \\
\leq& \underset{\mathbf{z}_m}{\mathbb{E}}\left\{ \left(1+\frac{2}{\sqrt{2\pi \left| \mathbb{T} \right|}-2}\right)\underset{\boldsymbol{\varepsilon}}{\mathbb{E}}\left[ \sup_{\mathcal{X},\mathcal{Y}\in \mathcal{H}}\frac{2}{\left| \mathbb{T} \right|} \boldsymbol{\varepsilon}^{\mathrm{T}}l_{\mathbb{T}}\left(\mathbf{z}_m,\mathbf{t}_m\right) \right] \right\} \\
\leq& \underset{\mathbf{z}_m}{\mathbb{E}}\left\{ \Lambda\left(1+\frac{2}{\sqrt{2\pi \left| \mathbb{T} \right|}-2}\right)\frac{2}{\left| \mathbb{T} \right|}\underset{\boldsymbol{\varepsilon}}{\mathbb{E}}\left[ \sup_{\mathcal{X},\mathcal{Y}\in \mathcal{H}}\boldsymbol{\varepsilon}^{\mathrm{T}}\mathbf{z}_m \right] \right\} \\
\leq& \Lambda\left(1+\frac{2}{\sqrt{2\pi \left| \mathbb{T} \right|}-2}\right)\frac{2}{\left| \mathbb{T} \right|}\underset{\boldsymbol{\varepsilon},\mathbf{z}_m}{\mathbb{E}}\left[ \sup_{\mathcal{X},\mathcal{Y}\in \mathcal{H}} \left\| \boldsymbol{\varepsilon} \right\|_{\mathrm{F}} \left\| \mathbf{z}_m \right\|_{\mathrm{F}} \right] \\
=& \Lambda\left(1+\frac{2}{\sqrt{2\pi \left| \mathbb{T} \right|}-2}\right)\frac{2}{\sqrt{\left| \mathbb{T} \right|}} \underset{\mathbf{z}_m}{\mathbb{E}}\left[ \sup_{\mathcal{X},\mathcal{Y}\in \mathcal{H}} \left\| \mathbf{z}_m \right\|_{\mathrm{F}} \right] \\
\end{align*}
\begin{align*}
=& \Lambda\left(1+\frac{2}{\sqrt{2\pi \left| \mathbb{T} \right|}-2}\right)\frac{2}{\sqrt{\left| \mathbb{T} \right|}} \frac{\mathbb{E}}{m+1}\sum^m_{i=0}\frac{1}{{\left| \mathbb{T}_1\cup \mathbb{S}_1 \right| \choose i}{\left| \mathbb{T}_2\cup \mathbb{S}_2 \right| \choose m-i}} \\
& \sum_{\mathbf{z}_m^{\left(1\right)}\subseteq  \mathbb{T}_1\cup \mathbb{S}_1}\sum_{\mathbf{z}_m^{\left(2\right)}\subseteq  \mathbb{T}_2\cup \mathbb{S}_2} \sqrt{\sum_{x_j\in \mathbf{z}_m^{\left(1\right)}}x_j^2+\sum_{x_j\in \mathbf{z}_m^{\left(2\right)}}y_k^2} \\
\leq& \Lambda\left(1+\frac{2}{\sqrt{2\pi \left| \mathbb{T} \right|}-2}\right)\frac{2}{\sqrt{\left| \mathbb{T} \right|}} \frac{1}{m+1} \\
& \mathbb{E}\sum^m_{i=0}\sqrt{\frac{{\left| \mathbb{T}_1\cup \mathbb{S}_1 \right|-1 \choose i-1}}{{\left| \mathbb{T}_1\cup \mathbb{S}_1 \right| \choose i}}\left\| \mathbf{x}_{\mathbb{T}_1\cup \mathbb{S}_1} \right\|^2_2+\frac{{\left| \mathbb{T}_2\cup \mathbb{S}_2 \right|-1 \choose m-i-1}}{{\left| \mathbb{T}_2\cup \mathbb{S}_2 \right| \choose m-i}}\left\| \mathbf{y}_{\mathbb{T}_2\cup \mathbb{S}_2} \right\|^2_2} \\
=& \sqrt{2}\Lambda\left(1+\frac{2}{\sqrt{2\pi \left| \mathbb{T} \right|}-2}\right)\mathbb{E}\sqrt{\frac{\left\| \mathbf{x}_{\mathbb{T}_1\cup \mathbb{S}_1} \right\|^2_2}{\left| \mathbb{T}_1\cup \mathbb{S}_1 \right|}+\frac{\left\| \mathbf{y}_{\mathbb{T}_2\cup \mathbb{S}_2} \right\|^2_2}{\left| \mathbb{T}_2\cup \mathbb{S}_2 \right|}} \\
\leq& \Lambda\left(1+\frac{2}{\sqrt{2\pi \left| \mathbb{T} \right|}-2}\right)\mathbb{E}\sqrt{\frac{\left\| \mathcal{X} \right\|^2_{\mathrm{F}}}{\left| \mathbb{T}_1 \right|}+\frac{\left\| \mathcal{Y} \right\|^2_{\mathrm{F}}}{\left| \mathbb{T}_2 \right|}},
\end{align*}
where the first inequality follows from the Theorem 3 in \cite{tolstikhin2015permutational}, the second inequality is a result of Rademacher contraction, the third inequality comes from the Hölder's inequality, the forth inequality is a consequence of arithmetic mean-quadratic mean inequality. Due to the hypothesis $\mathcal{H}$, the final bounds  can be derived by plugging (\ref{bound1}) and (\ref{bound2}) with $\alpha=1/\left| \mathbb{T}_1 \right|$ and $\beta=1/\left| \mathbb{T}_2 \right|$.

%
%

\ifCLASSOPTIONcaptionsoff
  \newpage
\fi



%
\bibliographystyle{ieeetr}
\bibliography{references_CTR-ALS}

%








\end{document}